\documentclass[conference]{IEEEtran}
\ifCLASSINFOpdf
\else
\fi
 \usepackage{multirow}
 \usepackage{pifont}
 \usepackage{xspace}
\usepackage{pgfplots}
\pgfplotsset{compat=1.17} 
\usepackage{tikz}
\usepackage[pass]{geometry}
\usepackage[normalem]{ulem}

\usepackage{epsfig,xspace,xspace,color,multirow}
\usepackage{subfigure,graphicx,amssymb,amsmath,endnotes,algorithm,algorithmic,caption}
\usepackage{subcaption}
\usepackage{listings}
\usepackage[hyphens]{url}
\usepackage[draft]{hyperref}
\usepackage{breakurl}
\usepackage{diagbox}
\usepackage{booktabs}
\usepackage[english]{babel}
\usepackage[utf8]{inputenc}
\usepackage{fancyhdr}
\usepackage{array}
\usepackage{listings} 
\usepackage{xcolor} 
\usepackage{tabularx}
\usepackage{soul}
\usepackage{hyperref}
\hypersetup{
  colorlinks,
  linkcolor={green!80!black},
  citecolor={red!70!black},
  urlcolor={blue!70!black},
	pdftitle={},
	pdfauthor={},
	pdfsubject={},
	pdfkeywords={},
	bookmarksnumbered=true,
	bookmarksopen=false,
	colorlinks=true,
	pdfstartview=FitB,
	pdfpagemode=UseOutlines
}

\definecolor{mypink1}{rgb}{0.858, 0.188, 0.478}
\newcommand{\cut}[1]{}

\usepackage{ifthen}
\newif\ifprintcomments

\newcommand{\mytodogrey}[1]{\textcolor{cyan}{\ding{46}~{\sf}~#1}}

\printcommentstrue

\ifprintcomments
    \newcommand{\zheng}[1]{}
    \newcommand{\revision}[1]{\textcolor{blue}{[#1]}}
    \newcommand{\hang}[1]{\textcolor{orange}{[Hang: #1]}}
    \newcommand{\zhiyun}[1]{}
    \newcommand{\haonan}[1]{\mytodogrey{[haonan: #1]}}
    \newcommand{\xingyu}[1]{}
    \newcommand{\zhiyune}[1]{}
    \cut{

    \newcommand{\zhiyun}[1]{\textcolor{cyan}{[Zhiyun: #1]}}
    \newcommand{\zhiyune}[1]{\textcolor{cyan}{#1}}

    \newcommand{\xingyu}[1]{\textcolor{teal}{[xingyu: #1]}}
    \newcommand{\mytodogrey}[1]{\textcolor{teal}{\ding{46}~{\sf}~#1}}
\else
    \newcommand{\zheng}[1]{}
    \newcommand{\hang}[1]{}
    \newcommand{\zhiyun}[1]{}
    \newcommand{\zhiyune}[1]{}
    \newcommand{\haonan}[1]{}
    \newcommand{\mytodogrey}[1]{}
    }
\fi

\newcommand{\sys}{\mbox{\textsc{LLMBisect}}\xspace}
\newcommand{\etc}{\emph{etc.}\xspace}
\newcommand{\ie}{\emph{i.e.,}\xspace}
\newcommand{\eg}{\emph{e.g.,}\xspace}

\newcommand{\squishlist}{
   \begin{list}{$\bullet$}
    { \setlength{\itemsep}{0pt}      \setlength{\parsep}{3pt}
      \setlength{\topsep}{3pt}       \setlength{\partopsep}{0pt}
      \setlength{\leftmargin}{1.0em} \setlength{\labelwidth}{1em}
      \setlength{\labelsep}{0.5em} } }
\newcommand{\squishend}{
    \end{list}  }    

\usepackage{xstring}
\newcommand{\PP}[1]{
\vspace{2px}
\noindent{\bf \IfEndWith{#1}{.}{#1}{#1.}}
}

\newcommand{\oldstuff}[1]{}
\definecolor{gray}{rgb}{0.5,0.5,0.5}

\begin{document}
%
\title{LLMBisect: Breaking Barriers in Bug Bisection with A Comparative Analysis Pipeline}


\author{%
\IEEEauthorblockN{%
Zheng Zhang \textsuperscript{\textasteriskcentered},
Haonan Li \textsuperscript{\textasteriskcentered},
Xingyu Li\textsuperscript{\textasteriskcentered},
Hang Zhang\textsuperscript{\textdagger},
Zhiyun Qian\textsuperscript{\textasteriskcentered},
}
\IEEEauthorblockA{\itshape\small
\makebox[\linewidth][c]{%
\textsuperscript{\textasteriskcentered}\,University of California, Riverside \enspace
\textsuperscript{\textdagger}\,Indiana University Bloomington \enspace
}}
\IEEEauthorblockA{\itshape\footnotesize
\makebox[\linewidth][c]{%
\textsuperscript{\textasteriskcentered}\,\{zzhan173, hli333, xli399, zhiyun.qian,krish\}@ucr.edu%
}\\
\makebox[\linewidth][c]{%
\textsuperscript{\textdagger}\,hz64@iu.edu \enspace
}}
}

\maketitle



\begin{abstract}
Bug bisection has been an important security task that aims to understand the range of software versions impacted by a bug, i.e., identifying the commit that introduced the bug. 
However, traditional patch-based bisection methods are faced with several significant barriers:
For example, they assume that the bug-inducing commit (BIC) and the patch commit modify the same functions, which is not always true.
They often rely solely on code changes, while the commit message frequently contains a wealth of vulnerability-related information.
They are also based on simple heuristics (e.g., assuming the BIC initializes lines deleted in the patch) and lack any logical analysis of the vulnerability.

In this paper, we make the observation that Large Language Models (LLMs) are well-positioned to break the barriers of existing solutions, e.g., comprehend both textual data and code in patches and commits.
Unlike previous BIC identification approaches, which yield poor results, we propose a comprehensive multi-stage pipeline that leverages LLMs to: (1) fully utilize patch information, (2) compare multiple candidate commits in context, and (3) progressively narrow down the candidates through a series of down-selection steps.
In our evaluation, we demonstrate that our approach achieves significantly better accuracy than the state-of-the-art solution by more than 38\%. Our results further confirm that the comprehensive multi-stage pipeline is essential, as it improves accuracy by 60\% over a baseline LLM-based bisection method.

\end{abstract}

\section{Introduction}
\label{sec:introduction}

N-day vulnerabilities are known security flaws that are often not fixed in a timely manner
due to complex dependency chains and limited maintenance resources \cite{elbaz2020fighting}.
The widespread reuse of open-source projects exacerbates this problem,
as there are usually multiple downstream distributions maintained by different parties in the ecosystem \cite{zhang2021investigation},
making it difficult to apply upstream security patches on time across all distributions.
Research has shown the popularity and severity of this problem in critical open-source projects such as Linux~\cite{li2024investigation} and Android~\cite{zhang2021investigation},
potentially affecting billions of users.


The information of affected software versions of a specific vulnerability is crucial for N-day vulnerability mitigation.
To obtain such information, it is necessary to locate the commit introducing the vulnerability (\ie bug-inducing commit, or \emph{BIC}) --- an essential task known as \emph{bug bisection}.
In this paper, we align with previous studies \cite{zhang2024symbisect} and define a \emph{BIC} as a commit that introduces a software bug into a program. It is possible for multiple commits  to contribute to the bug, with the final commit making the bug triggerable. In such cases, we consider the final commit as the \emph{BIC}, as it marks the point when the vulnerability is considered to exist.

Automated bug bisection can significantly speed up the bug-fixing process in downstreams (\eg 2.23x on average for Google's codebase, according to a previous study [7]),
however, achieving a high accuracy remains challenging.
Consequently, public information of vulnerable versions (\eg in NVD database \cite{nvd}) is usually incomplete or inaccurate, as shown in previous studies~\cite{anwar2021cleaning, wunder2024nvd, jiang2021evaluating}.


Existing automatic bug bisection approaches can be classified into several categories,
each with its own significant limitations:


\noindent\textit{(1) PoC-Based.}
Directly or symbolically execute the PoC (Proof-of-Concept) against each software version to test whether the vulnerability can be triggered. \cite{syzbot, zhang2024symbisect}
Though straightforward, this approach suffers from limited availability of vulnerability PoCs.
Furthermore, direct PoC execution~\cite{syzbot} often fails due to subtle variations across software versions,
resulting in low accuracy~\cite{syzbot_bisection_motivation},
while symbolic analysis \cite{zhang2024symbisect} is known to be expensive, and only supports limited bug types (\eg use-after-free, out-of-bounds memory access).


\noindent\textit{(2) Bug Report-Based.}
These approaches first collect available bug reports and then identify possible BICs by their ``relevance to the bug''~\cite{an2023fonte,wen2016locus,bhagwan2018orca},
with simplified assumptions such as ``a BIC should touch the code where the failure happens''.
Similar to PoCs, detailed bug reports are often not available.
Moreover, the simplified assumptions/heuristics may not hold in reality, reducing the accuracy, \eg Fonte's~\cite{an2023fonte} accuracy drops to 36\% when N=1 in its top-N ranking algorithm.

\noindent\textit{(3) Patch-Based.}
Being the most widely used,
these approaches statically analyze the bug-fix commit (usually available for vulnerabilities in open-source projects) to ``infer'' the bug-inducing commit in the commit history.
Existing techniques in this category~\cite{SZZ,vszz,VUDDY,V0finder,zou2017scvd,MOVERY} generally rely on manually developed, hardcoded, and thus inherently imprecise heuristics.
For example, one common one is to treat the commit that introduces one or more lines deleted by the bug-fix as the bug-inducing commit;
however, there are many situations where the bug-inducing commit and the fix commit do not intersect. For example, a patch can add an additional security check before the original vulnerable code (without removing any existing lines of code), potentially in a different function, or the deleted lines in the bug-fix are irrelevant.
Another significant shortcoming is that existing approaches usually only analyze the structured code changes in the bug-fix patches, while unable to take any advantage of the commit messages in the unstructured natural language form.
However, commit messages often contain rich and valuable information that can boost the bug bisection performance (\eg hints on the vulnerability root causes).



In this paper, we target patch-based bisection because it is the most widely applicable scenario --- not all bugs come with PoCs or crash reports.
We specifically have three goals:
i) support all types of patches and vulnerabilities,
ii) utilize full patch information including both code changes and commit messages, and
iii) go beyond the simple hardcoded heuristics and make accurate decisions based on analysis of the vulnerability logic.
To achieve these goals, we propose \sys, an LLM-powered highly accurate bug-bisection solution.
Our core insight is that LLMs are capable of understanding both code and natural languages,
extracting useful information for bug-bisection.
Recent LLM models (\eg OpenAI o1) also show impressive abilities in code reasoning.
Though promising, we find several obstacles to the direct application of LLMs. First, LLMs tend to produce \textit{excessive false positives}, aggressively and incorrectly labeling commits as bug-inducing. Second, LLMs suffer from \textit{self-consistency} issues, yielding conflicting decisions across multiple runs. 
Finally, the \textit{cost} of using LLMs on large-scale software is prohibitive: modern projects often contain tens of thousands of commits, and processing each one naively can lead to substantial token consumption and increased time and costs.

On the positive side, we observe that LLMs excel at comparative assessment, accurately selecting the true BIC from a small pool of candidates. To exploit this strength, we design a multi-step filtering pipeline that leverages LLMs’ comparative reasoning ability. First, we perform coarse-grained filtering to extract candidate BICs at scale (\S\ref{sec:candidate-generation}), using lightweight heuristics such as code changes and commit-message keywords. This step is inexpensive and efficient, yet it substantially narrows the search space for subsequent stages. Next, we apply fine-grained filtering in a multi-round, comparative fashion, ensuring that the LLM evaluates all promising candidates side by side. This design markedly improves accuracy. Finally, we incorporate majority voting at selected key points to mitigate self-consistency issues while avoiding significant performance overhead.

We extensively evaluate \sys on the Linux kernel, one of the most complex and important open-source software. The results show that \sys achieves a remarkable accuracy of 91\%, significantly outperforming state-of-the-art bug-bisection approaches. 

We summarize our contributions as follows:


\noindent\textit{(1)} We analyzed and articulated the limitations of existing bug-inducing commit (BIC) identification methods and proposed a novel and fundamentally different solution: an automated bisection tool based on LLMs, called \sys. Our approach incorporates previously overlooked information, i.e., commit messages, which often contain valuable cues about relationships to bug-inducing commits.

\noindent\textit{(2)}
We identified key challenges in directly applying LLMs to the bisection task and proposed a new multi-step filtering framework to address them. This design hinges on generating large numbers of candidate bug-inducing commits and leveraging the comparative power of LLMs to later prune them. 
We open-source our solution to facilitate further research \cite{LLMBisect}.

\noindent\textit{(3)}
We evaluated \sys against state-of-the-art methods and demonstrated that it significantly outperforms both existing tools and a baseline LLM-based solution. We also analyzed the root causes that limited the accuracy of earlier approaches and explained how our design overcomes these challenges. Through a comprehensive ablation study, we validated the effectiveness of our design across all stages.

\section{Motivation}
\label{sec:motivation}


\subsection{Motivating example}

\begin{figure}[h]
  \centering
  \includegraphics[width=\linewidth]{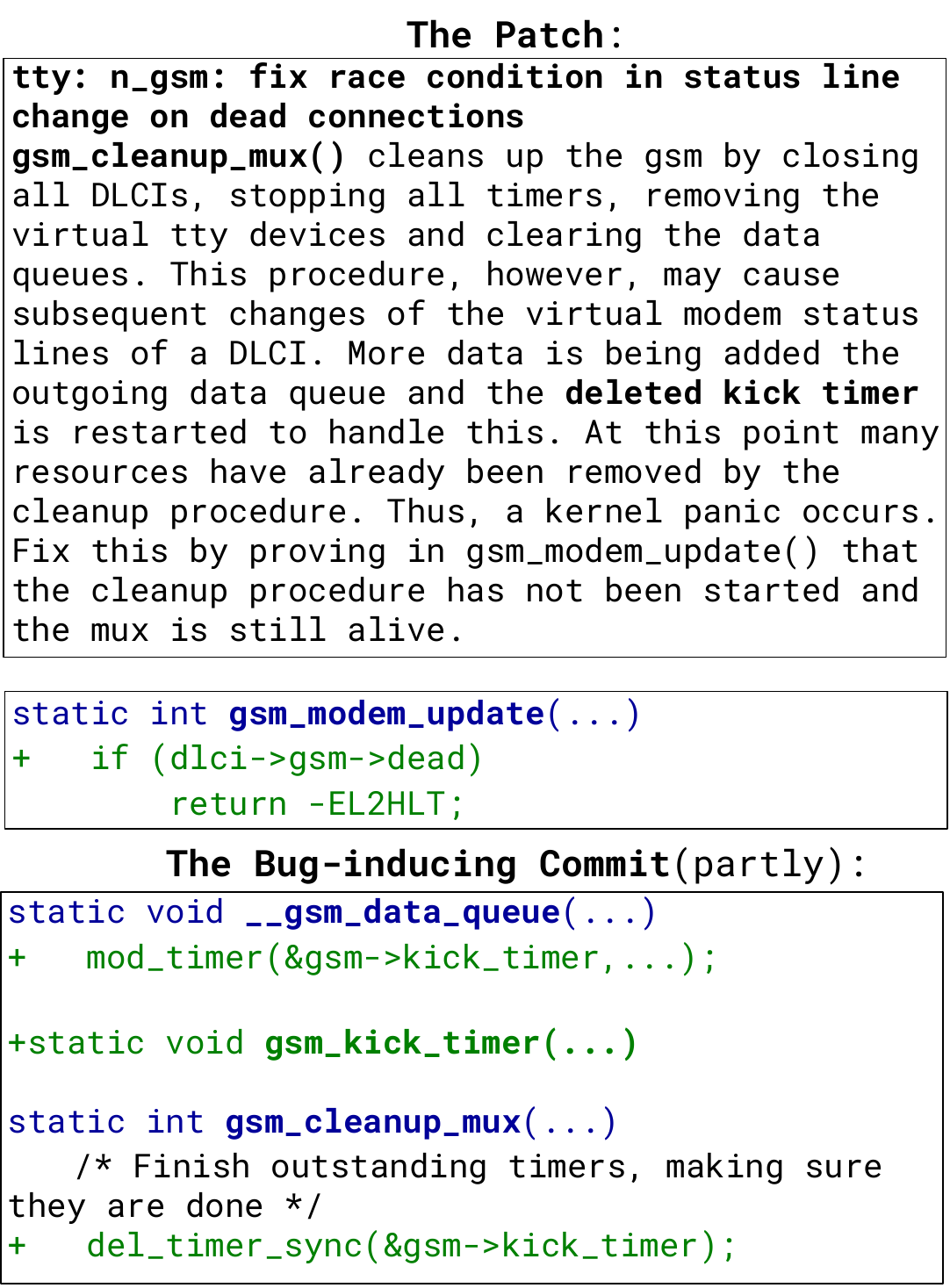}
  \caption{Motivating example}
  \label{fig:motivating example}
\end{figure}

\begin{figure}[h]
  \centering
  \includegraphics[width=\linewidth]{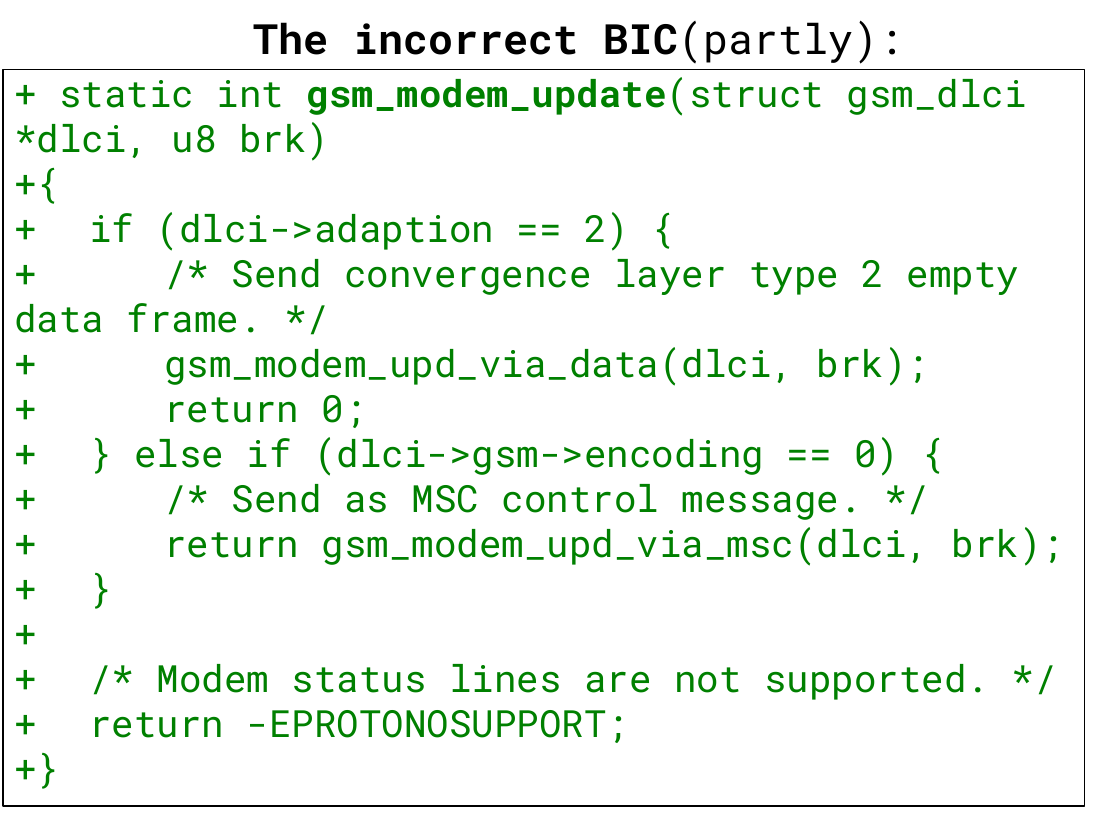}
  \caption{Motivating example FP}
  \label{fig:motivating example FP}
\end{figure}


Figure \ref{fig:motivating example} illustrates a race-condition-induced use-after-free vulnerability. However, in this example, the true BIC that introduced the vulnerability modified a completely different function from the one targeted by the patch.
Specifically, the \textbf{bug-inducing commit} introduced a new thread that runs the newly introduced function \texttt{gsm\_kick\_timer()}, while the \textbf{patch commit} adds a check in \texttt{gsm\_modem\_update()} to ensure that the cleanup process has not been initiated before it it allowed to create the timer thread (it will eventually call \texttt{\_\_gsm\_data\_queue()}), and hence eliminating the possibility of a race. 

\vspace{2pt}
\noindent
\textbf{SymBisect}, the state-of-the-art PoC-based Bisection method, cannot accurately extract the BIC in this case because:
1.	There is no existing Proof of Concept (PoC).
2.	SymBisect only supports two specific types of bugs: Out-of-Bounds (OOB) and Use-After-Free (UAF). Specifically, it does not support race condition cases.

\vspace{2pt}
\noindent
\textbf{VSZZ}, the state-of-the-art method in the SZZ family~\cite{SZZ,AG-SZZ,MA-SZZ,RA-SZZ,vszz}, fails to handle such cases because its fundamental assumption is that the BIC modifies the same functions as the patch (specifically, that the BIC initializes the lines deleted in the patch). However, in this case, the BIC and the patch modify completely different functions.

\vspace{2pt}
\noindent
\textbf{V0Finder}, an advanced vulnerable code clone detection method, identifies vulnerable versions by comparing the patch functions of the target version and the patch functions before the patch, after normalization and abstraction. If they are identical, the target version is considered vulnerable. However, this method fails in the illustrated case because the BIC does not modify the patch functions at all.

This case motivates us to think of a better approach for extracting candidate commits, one that goes beyond merely tracking patch functions. In fact, in this example, although the BIC modifies a different function than the patch, we note that \texttt{gsm\_cleanup\_mux()} is explicitly mentioned in the patch description. This provides an important hint that we can expand our focus to not only analyze code changes but also extract valuable information from commit messages.



\begin{figure}[h]
  \centering
  \includegraphics[width=\linewidth]{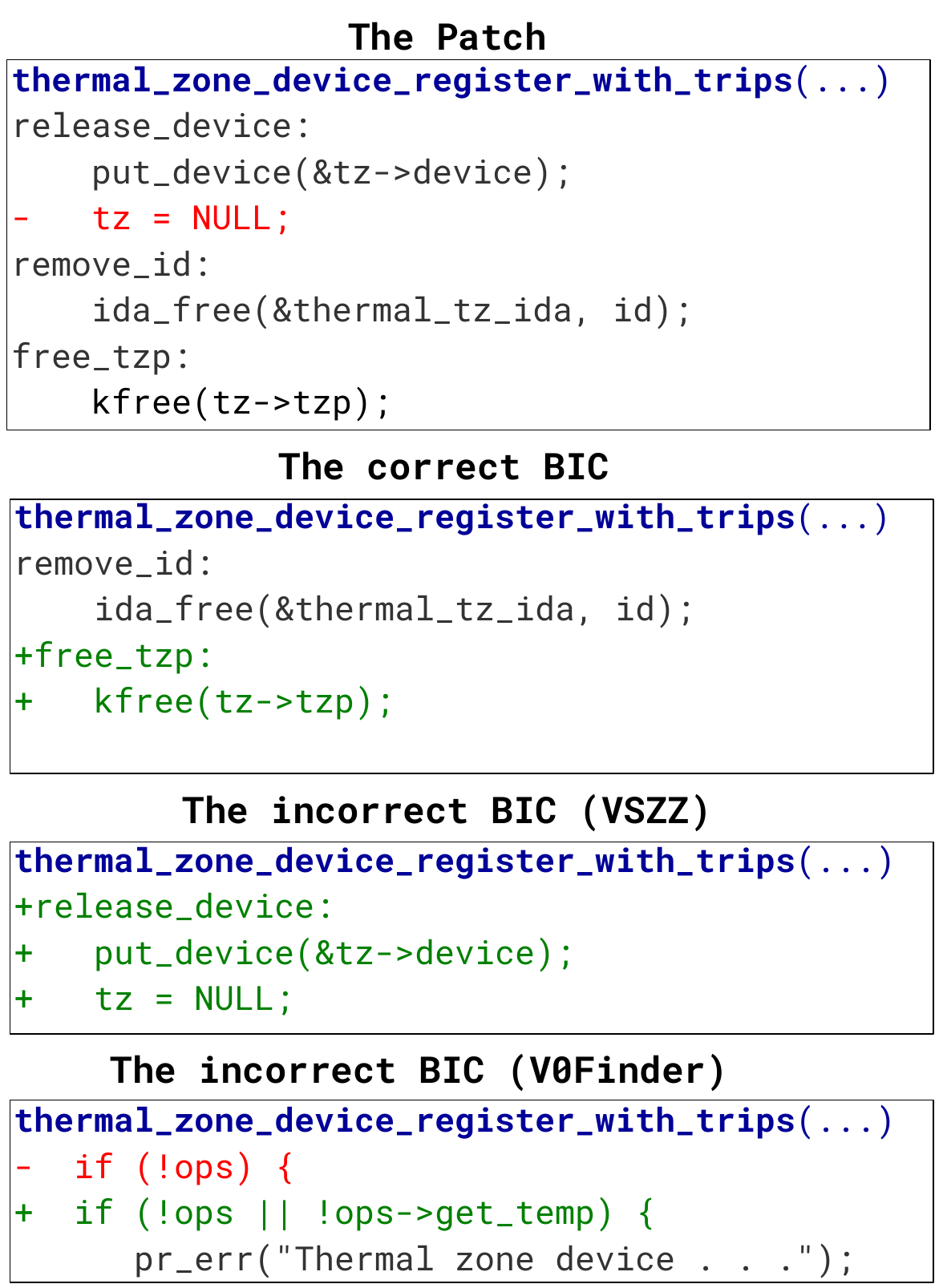}
  \caption{Motivating example \#2}
  \label{fig:motivating example2}
\end{figure}

Figure \ref{fig:motivating example2} illustrates another motivating example, which represents a NULL pointer deference bug. In this example, although the BIC modifies the same function as the patch and is thus included among the candidates, the flawed heuristics of traditional methods prevent them from accurately identifying the correct BIC. 

Specifically, the buggy code incorrectly sets \texttt{tz} to NULL 
under certain conditions, which causes the NULL pointer to be dereferenced subsequently in \texttt{kfree(tz->tzp)}.
The patch fixes this vulnerability by removing the assignment that sets tz to NULL.

The commit introducing this vulnerability added a \texttt{kfree()} function call where the null dereference occurs. Before the BIC, the \texttt{kfree} function call did not exist, so naturally, the null dereference was not an issue.

\textbf{VSZZ} does not try to understand the logic of the vulnerability. Instead, it tracks the lines deleted in the patch, leading back to the commit that initialized the line (in this case, an earlier commit that first created the line of \texttt{tz = NULL}). This completely overlooks the actual BIC.



\textbf{V0Finder}’s flawed heuristics, comparing hash values (essentially string matching after normalization and abstraction) of the whole patch function,   take a different approach, focusing on all commits that modified the patched function. Specifically, it identifies a commit that modified the patch function (the latest one before the patch) as the BIC, but this modification is unrelated to the vulnerability. V0Finder does not determine whether the modification is logically connected to the vulnerability; it simply assumes that, before the modification, the function was different, and therefore, the vulnerability did not exist. 


\subsection{Limitations of previous methods}


Based on the motivating examples, we summarize the key weaknesses in patch-based methods, including \emph{SZZ algorithm and its variants}~\cite{SZZ,AG-SZZ,MA-SZZ,RA-SZZ,vszz} and most vulnerable code clone detection solutions~\cite{VUDDY,V0finder,zou2017scvd,MOVERY}.
They suffer from the following limitations:

1) They often only consider code changes, ignoring commit messages, which frequently contain crucial information about vulnerabilities. 

2) They fail to account for cases where a Bug-Introducing Commit (BIC) does not change the functions affected by the patch.

3) Many of them focus on deleted lines in the patch, making them ineffective when patches only include added lines or when the deleted lines are not critical to the vulnerability.

4) They tend to treat all code changes (such as deleted lines) equally. In reality, not all changes are of equal significance to the vulnerability.

5) Their judgments are often based on simple heuristics rather than logical reasoning. For example, VSZZ, the state-of-the-art SZZ method, traces back commit history to the earliest commit (instead of the most recent) that introduces the deleted lines of a patch. Such heuristics are often not accurate.

\subsection{Insights}

Revisiting the motivating example, we propose three design goals for an improved solution:

	1)	Leverage Full Patch Context: The solution should utilize the complete patch context, including both the patch code diff and commit messages, as these provide critical clues about the bug-inducing commit.
    
	2)	Minimize Assumptions and Requirements: Unlike approaches such as VSZZ, the solution should support patches that only add lines. It should also handle all types of bugs rather than being restricted to specific categories (e.g., SymBisect). Additionally, it must accommodate patches that do not modify functions, a limitation seen in V0Finder.
    
	3)	Incorporate Logical Reasoning: The approach should analyze the logic of the vulnerability to make a decision on the bug-inducing commit, rather than depend on simplistic and hardcoded heuristics like those used in VSZZ and V0Finder.

To achieve these goals, we propose leveraging large language models (LLMs) for the task of bug bisection. LLMs are well-suited for this purpose due to their ability to comprehend both code and patch descriptions. Moreover, they are trained on all types of bugs and patches and thus not limited to reasoning about specific types of bugs/patches. LLMs have demonstrated effectiveness in various bug analysis tasks~\cite{xu_large_2024, wang_sanitizing_2024, li2024enhancing, wang_llmsa_2024}
and have been improving one generation after another.

\section{Design}
\subsection{Design Motivation}
\label{sec:design-motivation}



Though LLMs show great potential in enhancing existing bug bisection techniques, it remains unclear how to best leverage their capabilities for optimal performance. To explore this, we began by reproducing the typical workflow of previous bug-inducing commit (BIC) identification work with LLM's drop-in help, which serves as the baseline for our design.

\noindent\textbf{A baseline LLM-based bisection method.}
The method is inspired by heuristics employed in classic bug bisection methods (\eg SZZ). It operates as follows: 1) lists the commits that modified the patch function -- this is an extended set of candidate BIC commits compared to prior methods like SZZ, and 2) queries the LLM for each commit in the list in reverse chronological order, stopping at the first one identified as the BIC. Through this baseline and its subsequent variants, we encountered several challenges that led to a notably low accuracy — the baselin achieves only 30\%, as will be shown in \S\ref{sec:eval}. These challenges revealed key limitations of applying LLMs without structural guidance in the bug bisection task. They also led to new opportunities for improvement, enabled by a better understanding of BIC characteristics and the strengths and weaknesses of LLMs in the bug bisection task. These insights directly informed our design choices and gradually shaped the final version of our design through multiple iterations.

Intuitively, bug bisection is the process of identifying the BIC from a list of candidate commits related to a given patch and its associated vulnerability.
Thus, we can divide the process into two steps: 
1) extracting candidate commits from the commit history, and 2) selecting the exact bug-inducing commit from the candidate commits. 




\subsubsection{Collection of BIC candidates}

\textbf{Baseline.} 
Our aforementioned baseline method generalizes the state-of-the-art method, i.e., VSZZ and V0Finder, which considered only the commits that changed patch function(s).
Specifically, the baseline method collects \emph{all historical commits that modify the patched function(s)}.
The intuition is that this represents a superset of commits encompassing the BICs identifiable by previous methods. It can also overcome their limitation of not supporting patches with added lines only (no deleted lines). 

\noindent\textbf{Challenge \#1.}
The total number of commits that modified the patch function is often quite large in the commit history. Too many candidates can reduce the accuracy of the LLM (as observed in our preliminary experiments). 
Moreover, some real-world BICs do not modify the patch functions at all, which will be missed by the above solution.

\noindent\textbf{Observation \#1.}
Not all functions or lines modified in the bug-fix commit are equally important or relevant to the vulnerability. For example, some code changes are merely for refactoring purposes without changing the semantics of the code. Previous methods also attempt to identify irrelevant code changes. However, their methods are limited to only simple patterns such as adding or removing comments \cite{V0finder, VUDDY}.

\noindent\textbf{Solution \#1.}
We change the simple, non-distinguishing function-based candidate selection to a fine-grained, critical-line-based selection.
Specifically, we first identify the most relevant changed lines to the vulnerability from the bug-fix commit, with LLM's help, and then include only historical commits that touch these lines in the candidate list.
Note that this method is no longer limited to changes within the patched functions. Instead, it also considers changes to global variable definitions and struct definitions as potentially critical. As a result, this approach not only significantly reduces the number of candidates to inspect (by 81\% on average in our evaluation) but also enables the identification of new, previously overlooked candidates.

It is worth noting that we also considered further expanding the scope for critical line selection (e.g., callers of the patched functions). However, this will substantially bloat the number of BIC candidates, with little benefit. As will be shown in \S\ref{sec:eval-acc}, there are 8 cases where the BIC modified files are completely different from those in the patch, causing the BIC to be excluded from our candidate set. None of these cases could be resolved by including the callers of the patched functions in the analysis.


\noindent\textbf{Challenge \#2.}
While this improves the accuracy if the BIC indeed modified the critical lines, it still does not solve an aforementioned problem --- the code change made in the bug-inducing and bug-fix commits can be disjoint (\eg in different functions or files).

\noindent\textbf{Observation \#2.}
The patch commit messages often contain useful clues hinting at the vulnerability's root cause and connecting it to the bug-fix (\eg the commit message of a bug-fix in \texttt{foo()} may mention that the vulnerability originates from \texttt{bar()}). The motivating example illustrated this point. 

\noindent\textbf{Solution \#2.}
Going beyond the function- and critical-line-based candidate selection,
we can leverage LLMs to select additional BIC candidates using hints extracted from the commit messages (\eg commits that modified a function mentioned in the commit message).
Because we look for functions or variables outside of the patched function, it is complementary to the previous two methods by design.


\subsubsection{Selection of BIC from candidates}

\textbf{Baseline.} As mentioned in the aforementioned solution, we follow SZZ-style bisection, which simply inspects the BIC candidates in reverse chronological order; the first one recognized as BIC will be the selected one. This is a reasonable choice because we define a BIC as the last commit contributing to the vulnerability.
While plausible, we identified multiple challenges during our preliminary experiments.


\noindent\textbf{Challenge \#3.}
High false positive rate of LLMs.
During reverse chronological traversal, the LLM tends to identify BICs too eagerly, causing it to stop prematurely and miss the actual bug-inducing commits, which leads to low accuracy.

\noindent\textbf{Observation \#3.}
Despite having FPs, LLMs perform well in discerning the real BIC when it is presented with multiple candidates. The LLM’s strength in comparative reasoning eliminates the need to have them make definitive decisions about individual BIC candidates in isolation.

\noindent\textbf{Solution \#3.}
We adopt a two-round BIC selection:
1) let the LLM inspect \emph{all} candidates and identify \emph{all} potential BICs, without early termination, and
2) let the LLM \emph{compare} all the identified BICs and select a final one.


\noindent\textbf{Challenge \#4.} 
False negatives are incurred in using any single method to collect BICs.
Vulnerability-relevant lines can still sometimes be missed by LLMs in some cases, resulting in false negatives in BIC recognition.

\noindent\textbf{Observation \#4.}
The three methods of generating BIC candidates can complement each other in terms of the covered BIC candidates. 

\noindent\textbf{Solution \#4.}
To avoid missing the correct BICs, we feed all three sets of BIC candidates (generated by three different methods) to the LLM, using the three methods described earlier. Although this approach increases token consumption compared to using a single set, it helps improve coverage. To further improve accuracy, we make a final selection from the results generated by different methods (\eg function-based and critical-line-based), rather than merging the candidate commits at the beginning. This is because the accuracy of the LLM tends to drop when we feed a large set of BIC candidates. 
In other words, we will feed only three candidates to the LLM, when it is making the final verdict. 
In most cases, even if a method produces an incorrect candidate, it is unlikely to be selected among the final three. However, when it does generate the correct candidate, that candidate is very likely to be selected in the final stage. This is again taking advantage of the comparative reasoning power of the LLM.



\begin{figure}[ht]
  \centering
  \includegraphics[width=1\linewidth]{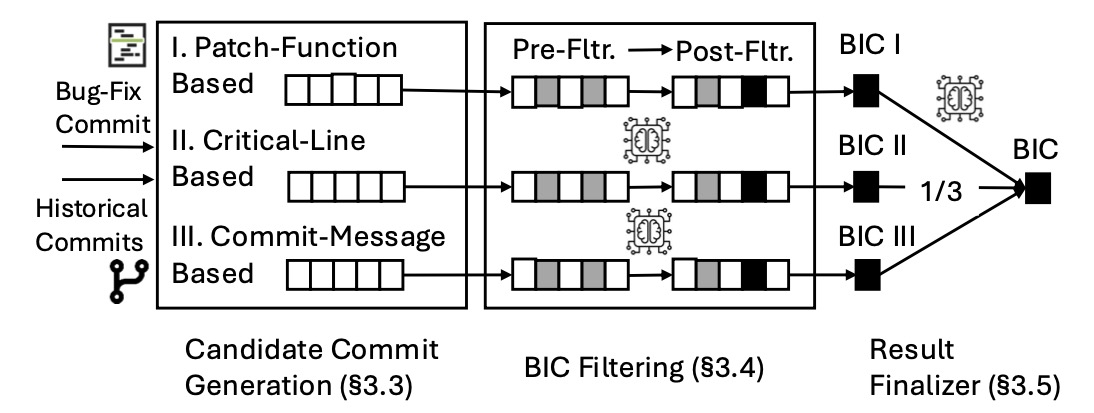}
  \caption{Workflow of \sys}
  \label{fig:design}
\end{figure}




\subsection{Workflow} 
\label{sec:design-workflow}

Motivated by the above design explorations, we present the workflow of our final design of \sys in Fig.~\ref{fig:design}.
As we can see, there are three overall stages: 1) Candidate Commit
Generation.  2) BIC Filtering. 3) Result
Finalizer.

\noindent\textbf{Candidate Commit
Generation.} Given a bug-fix commit,
this stage's goal is to list all historical commits that could potentially be the BIC for future investigation (\ie candidate generation). As described previously, we have three candidate commit generation methods, based on patch functions, critical lines, and commit messages, respectively. These methods can complement each other (Solution \#1 and Solution \#2 in Section \ref{sec:design-motivation}).

\noindent\textbf{BIC Filtering.} 
At this stage, we aim to select the most likely Bug-Inducing Commit (BIC) from each list generated in the first stage, resulting in up to three final BIC candidates. This process is divided into two phases: the pre-filtering phase, which identifies possible BICs, and the post-filtering phase, which selects the most likely BIC (Solution \#3 in Section \ref{sec:design-motivation}).

\noindent\textbf{Result Finalization.} At this stage, we finalize our decision by selecting one final Bug-Inducing Commit (BIC) from the potential BICs (up to three) identified during the BIC filtering stage. (Solution \#4 in Section \ref{sec:design-motivation})

\noindent\textbf{Majority voting.}
LLMs sometimes make different decisions regarding BIC selection in multiple runs (\ie self-consistency).
We observed that there usually exists a ``dominating'' decision occurring in most runs. Thus we adopt a ``majority voting'' mechanism in our design,
where we run LLMs multiple times (defaulted at 7) for BIC identification during the Result Finalization phase.

\subsection{Candidate Commit Generation}
\xingyu{it is better if we can provide prompt template.}
\label{sec:candidate-generation}

The quality of the candidate lists can significantly impact the accuracy of the final BIC identification.
On the one hand, too many commits in the list will simultaneously increase the likelihood of errors and the cost.
On the other hand, missing relevant commits leads to false negatives.
As a result, we would ideally like the list to (1) contain the true BIC, and (2) be small enough.
In practice, these two goals are hard to achieve simultaneously.
We will present our key design below to strike a good balance.


\PP{Function-based Generator.}
As used in the baseline method (\S\ref{sec:design-motivation}),
the most commonly used generator in existing work is based on patched function(s) in the bug-fix commit,
where all historical commits modifying the same function(s) are selected as candidates.
This strategy is effective as bug-inducing and -fix commits frequently modify the same function(s).


\PP{Critical-line-based Generator.}
\cut{
Though comprehensively capturing BIC candidates modifying the same function,
patch-based generator has two problems:
i) the function granularity is coarse for subtle security vulnerabilities --- a commit changing the same patched function may still be irrelevant to the vulnerability logic and introduction, and
ii) the vulnerability can be out of the function scope (\eg lies in a global variable definition),
whose BIC will be excluded from the candidate list by a function-based generator. 
}
First, it recognizes lines  that are truly relevant to the vulnerability logic (\ie critical lines).
To achieve this, we utilize LLM's ability to comprehend both code and natural language to recognize critical lines, which are far more accurate than heuristic-based approaches used in existing work. 
We also provide the LLM with the full definitions of the patched functions, as part of prompt engineering, to better facilitate its understanding of vulnerability logic.
Second, we will only treat historical commits that modify critical lines as BIC candidates. 


Conceptually, we would like an LLM to focus on particular parts of the code that pertain to the vulnerability, whether they are part of the patched functions or changes to a global variable definition (if it is included in the code diff). It turns out that it is a non-trivial task. As mentioned, prior work often relies on overly simplistic heuristics to define critical lines. For example, all deleted lines within a function are considered critical~\cite{vszz}, or every line in the patched functions is defined as critical~\cite{V0finder}.
We would like to generalize it and improve it, with the help of LLMs.
\zheng{updated}

In particular, we divide patches into three types and apply tailored strategies using LLMs to identify critical lines:


\noindent\textit{(1) Patches with deleted lines.}
Deleted lines in a bug-fix commit are often related to the vulnerability,
so in this case, we narrow our scope of critical line identification to the deleted lines (excluding trivial ones like comments) to improve efficiency.
However, if LLMs recognize no critical lines among those deleted,
we expand our scope to the whole patched function.

\noindent\textit{(2) Patches with only added lines.}
If a patch has only added lines, previous solutions, such as VSZZ simply give up. 
However, we would extract critical lines from the entire modified function/struct.
Specifically, we would feed the whole patch, including the code diff and commit message, as well as the complete definitions of the affected functions. For example, if the patch merely adds a range check for a variable (such as an array index), the LLM can analyze the commit message and the function’s surrounding code to identify critical statements related to this variable (e.g., an array access with the index, where the out-of-bounds (OOB) error occurs). These critical statements are often modified in the BIC.
\zhiyun{Can we use an example to illustrate the effectiveness of this? Perhaps an example where the commit message is clearly helping.}  \zheng{updated}

\noindent\textit{(3) Patches with only reordered lines.}
\zhiyun{How do we identify such patches? Using LLMs or static analysis?} \zheng{string match and a simple script are enough}
These patches merely change the line positions (\eg adjust the critical section length by moving the lock/unlock statements).
Here vulnerabilities are usually caused by improper relative positioning of two lines, one being the line modified by the patch and the other whose relative order to the modified line has changed. Therefore, merely focusing on the presence of modified lines is insufficient to determine whether a vulnerability exists. The introduction of a vulnerability is often closely related to the other line. \zhiyun{should give some more reasoning about why it is insufficient to just look at the changed lines.}
Therefore, for such patches, we extract critical lines from the modified lines and the affected context statements ( the statements whose relative position to the modified line has been altered after applying the patch). For example, if a patch moves a \texttt{lock()} call to an earlier position in the function, thereby extending the scope of the lock to include more statements, the statements newly encompassed by the lock after the patch are considered affected context statements. These often include critical statement related to the vulnerability.

\zhiyun{Can we show our LLM prompt template for these three cases? Our description is quite concise. I wonder whether there is anything interesting to show in the prompts, e.g., asking LLMs to pay attention to the commit message.} \zheng{updated}


\noindent\textbf{Commit-message-based Generator.} 
As discussed in \S\ref{sec:design-motivation}, neither of the above generators can correctly include the BIC candidate if it has no overlap (regarding the modified code) with the bug-fix commit.
To address this issue, we design the third generator based on commit messages,
from which we extract valuable information regarding the vulnerability. More specifically, we try to extract the following information from the commit message with regular expression matching:

\noindent\textit{(1) Function/struct/variable names.}
They could indicate the actual location of the vulnerable code or global variables.

\noindent\textit{(2) Commit hashes.}
Some commit messages directly reference earlier (BIC) commits by their hashes.

We also include names of modified functions by the bug-fix as keywords,
though technically they are not extracted from the bug-fix commit messages.
They are useful because even the BIC may not modify the same functions,
it might still modify their callers which contain their names in the code (e.g., adding a call to the patched function).   

To avoid redundant execution, we disregard all functions or structs modified by the Bug-fix commit (which have already been tracked by the first two generators).
\zhiyun{This part can use some more detail to describe. For example, some details in the implementation section can be moved here. At least we should tell people what kinds of information we try to extract. Also useful to show how the prompts look like.}\zheng{updated}






After running the three above candidate generators independently,
we obtain \emph{three} candidate lists at the end of this stage, which will be fed as input to the next stage (\S\ref{sec:bic-selector}).

\subsection{BIC Filtering}
\label{sec:bic-selector}

In \S\ref{sec:candidate-generation}, we generate three lists of candidates,
at this stage, we try to pick \emph{one} most likely BIC from \emph{each} list,
resulting in \emph{up to three} final BIC candidates (it is posible that no BIC is selected from a certain list) for the next stage (\S\ref{sec:result-merge}).
One straightforward way to pick the BIC from a candidate list, as mentioned in \S\ref{sec:design-motivation},
is to inspect each commit in reverse chronological order and stop when one is recognized as the BIC.
However, this leads to a high positive rate because the ``most likely'' BIC infers that it can only be reliably identified from a comparison of multiple potential ones.
Therefore, we design our BIC filtering process to be composed of two sub-phases: the pre- and post-filtering.


\noindent\textbf{(1) Pre-Filtering.}
For \emph{every} commit in a candidate list, we prompt it with the original bug-fix commit to LLM for a decision regarding whether it \emph{could} be the BIC.
This will result in multiple potential BICs selected by the LLM.

\noindent\textbf{(2) Post-Filtering.}
The LLM is then instructed to perform a \emph{comparative assessment} of all selected BICs in the pre-filtering phase,
to finally pick \emph{one} most likely BIC per candidate list.

This design gives LLMs sufficient opportunities to review all candidate commits and carefully compare them for better-informed decisions,
significantly boosting the BIC identification accuracy, compared to baseline early stop solution.







    

\subsection{Result Finalization}
\label{sec:result-merge}

\begin{figure}[ht]
  \centering
  \includegraphics[width=0.7\linewidth]{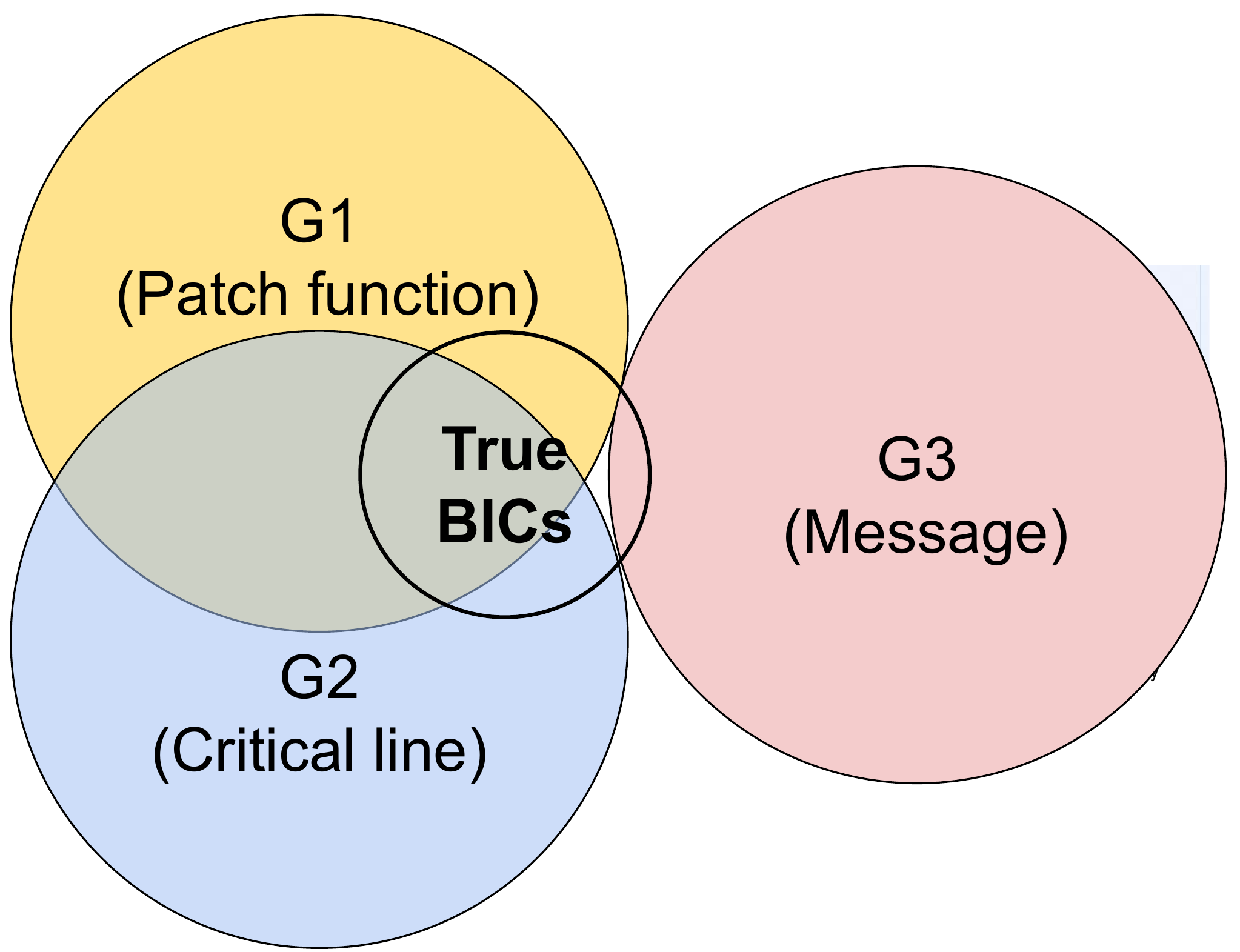}
  \caption{Candidate Generators}
  \label{fig:Venn}
\end{figure}

The last BIC filtering stage (\S\ref{sec:bic-selector}) outputs up to three potential BICs selected from multiple candidate lists,
while we still need to finalize our decision by picking one final BIC.
To achieve this, our procedure is similar to the last stage (\S\ref{sec:bic-selector}).
Specifically, we present all BIC candidates (up to three) to the LLM for a comparative evaluation,
in order to reach a final BIC decision as \sys's output.
Note that though rarely the case, it is possible that \sys eventually fails to output any BIC (\eg zero candidates were selected in the previous BIC filtering or this result finalization process).



As mentioned in \S\ref{sec:design-motivation}, we have three methods of generating BIC candidates. They can complement each other. They each have their trade-offs regarding the two goals listed above. 
Conceptually, we can see Fig.~\ref{fig:Venn} which illustrates the different candidate sets produced by different generators.
As a result, 
\sys adopts all three of the aforementioned BIC candidate generators.
Our design is to have them work independently initially.
Later on, we will attempt to pick the final result with Result Finalization. 
Note that we do not want to merge all the candidates into the same set initially and then have LLM pick one. This is because such a set will be too large which will hurt the accuracy. As mentioned in \S\ref{sec:design-motivation}, the number of candidates produced by the function-based generator is already large, limiting the LLM's accuracy in picking the right BIC. 
It is therefore beneficial to keep the set of candidates produced by critical-line-based and commit-message-based methods separate.
This way, if the correct BIC is located in either of the two sets, it will likely be correctly identified by the LLM. 
Again, the function-base generator can be viewed as a backup option. In case the correct BIC is present in only its result, then at least we would still have a chance to identify it.



\section{Implementation}

We implement a prototype of \sys with 5,331 LoC in Python.
In this section, we discuss some noteworthy implementation details.


\noindent\textbf{Function-Based Candidate Generation.}
One can use a git command to track all commits that modify a specific function:
\texttt{git log \textless commit\_hash\textgreater \, -L:\textless funcname\textgreater:\textless filename\textgreater}.
However, it can miss some commits when the function has been renamed or the file that contains the function has been renamed.
To address this limitation, we developed a script to track all commits that modify the given file/function more comprehensively, correctly handling the renaming issues.
For each commit, we then extract the functions modified by it, enabling us to obtain the complete list of commits modifying the specific function.

\noindent\textbf{Patch Type Classification.} 
We implement a Python script to first determine the type of patch (e.g., those with only added lines). This is relatively straightforward. We first extract and ignore all changes relating to comments, and then can easily classify patches into those with only added lines and those with deleted lines (we do not differentiate the patches with only deleted lines). 
Among the patches with deleted lines, if there are also added lines, we then use a simple string-match-based heuristic to identify reordered statements. Specifically, we consider a patch as reordering changes if and only if all the changes are related to reordering. In other words, all the removed lines must show up as added lines in another location verbatim.

After collecting this information, we first obtain historical commits that modify the same files as the bug-fix commit, then for each commit, we check whether it matches any of the extracted information with the commit message (\eg has the same hash, change/call the mentioned function, \etc).
If so, we also consider it as a BIC candidate, which can be missed by function-based or critical-line-based generators.
\noindent\textbf{LLM Models.} In our implementation, we primarily use OpenAI o1 (\texttt{o1-preview-2024-09-12}) as the main LLM. We also evaluate other models, including GPT-4o (\texttt{gpt-4o-2024-08-06}) and the open-source, Llama 3 (\texttt{nvidia/llama-3.1-nemotron-70b-instruct}). The results of these evaluations are presented in Section~\ref{sec:eval-ablation}. The specific prompts used in each step are included in the appendix.

\section{Evaluation}
\label{sec:eval}

In this section, we evaluate \sys to answer the following research questions:

\squishlist
\item RQ1: How accurately does \sys identify BICs?

\item RQ2: How does \sys compare against other state-of-the-art BIC identification methods? 

\item RQ3: How does each component and phase of the pipeline of \sys contribute to its final performance?

\item RQ4: How costly is the solution?
\squishend



\noindent\textbf{Dataset.}
We evaluate \sys against Linux kernel CVEs.
Several key considerations inform this choice:
(1) Linux kernel is one of the most important and widely used software, its ecosystem contains numerous downstream distributions potentially impacted by N-day vulnerabilities, highlighting the importance of an accurate bug bisection,
(2) The kernel also has one of the most complex codebases, containing a wide range of vulnerabilities reported daily by security practitioners.
We believe the diversity and complexity of Linux kernel CVEs can rigorously test \sys's accuracy and reliability.
Note that despite our choice, \sys by design is agnostic to the target software or vulnerability types.



Given the sheer number of Linux kernel CVEs and the high cost of advanced LLM tokens (e.g., o-1),
we randomly sampled 100 CVEs in each of 2023 and 2024 (200 in total).
We specifically include CVEs in 2024 as they are published after the LLM knowledge cut-off date,
validating whether \sys's result is influenced by the LLM's pre-existing knowledge about the CVEs (As shows later,  there is nearly no difference. \sys demonstrated similar accuracy on CVEs from 2023 and 2024).
We also included CVEs in 2023 because the CVE assignment criteria became more relaxed starting from 2024 (\eg many non-security issues also had CVEs assigned)~\cite{kernel-cna,kernel-cna1,may24cve}.\xingyu{add the reference, please check; also, I feel that "many non-security issues also had CVEs assigned" may be too strong, the reference does not show this. "the number of assigned CVEs are bloated, such as 1000+ assigned in May 2024, may be more suitable }
Testing these CVEs demonstrates \sys's accuracy on security vulnerabilities more reliably.


\noindent\textbf{Ground Truth.}
To get the ground truth (\ie the correct BIC for a specific vulnerability),
we intentionally include in our dataset only those CVEs whose fix commit has a \emph{fixes tag} \cite{fix_tag}, which points to the BIC(s) given by kernel developers.
We then manually verify them according to our BIC definition and assemble the ground truth. \zhiyun{can we give some examples in terms of how we fix the fixes tag? I remember reviewers asking about this for our SymBisect submission.} 
After manual verification, we identified and corrected 11 cases with inaccurate fix tags.
It is important to note that fix tags are merely used to provide the ground truth that is otherwise difficult to obtain -- \textit{we remove the fixes tags from bug-fix commits during the experiments}.


\noindent\textbf{Threats to Validity}.
In our sampling process, we followed the same approach used in prior studies (e.g., SymBisect), selecting commits that include ``Fixes" tags and manually verifying them. One potential bias in this approach is that commits with ``Fixes" tags may contain more detailed and descriptive commit messages, which can make them easier for LLMs to process and understand.
Unfortunately, there is a lack of dataset without ``Fixes'' tags to validate or invalidate the hypothesis.  
As a result, we choose to follow the best available approach used by prior methods. 
Nonetheless, we acknowledge the potential biases introduced by this approach and leave the task of building a reliable benchmark of bugs without ``Fixes" tags as future work.

\noindent\textbf{Comparison Targets}.
We extensively compare \sys with three state-of-the-art tools covering different bug bisection methodologies:

\noindent\textit{(1) PoC-based bisection.}
SymBisect\cite{zhang2024symbisect} is a state-of-the-art PoC-based bisection tool.
It generates various guidance from PoC execution traces and uses principally guided under-constrained symbolic execution to confirm the bug's existence.
However, it only supports limited vulnerability types and relies on PoCs --- unavailable for most Linux kernel CVEs.\xingyu{can we give a percent?}
Even then, we would like to see how \sys compares to SymBisect using its evaluation dataset, which SymBisect supports very well.
This is an interesting experiment that can showcase the performance differences between the symbolic reasoning (in SymBisect) and LLM's reasoning (in \sys).

\noindent\textit{(2) Patch-based bisection with SZZ-style algorithms.}
As mentioned in \S\ref{sec:motivation}, 
SZZ-style algorithms generally rely on the assumption that BIC will initialize lines deleted in the bug-fix commits.
We select VSZZ~\cite{vszz} --- the state-of-the-art open-source tool in this domain --- as a comparison target,
we configure it with default options specified in its tutorial~\cite{Vszz_opensource}.

\noindent\textit{(3) Patch-based bisection with vulnerable code clone detection.}
These methods are based on code similarity comparison between the vulnerable pre-fix version and a specific target version to probe the first vulnerable version (\S\ref{sec:motivation}).
V0Finder~\cite{V0finder} is a latest tool in this direction,
we configure it with its default options~\cite{V0Finder_opensource} in our comparison.

\begin{table}[]
\scalebox{0.8}{
{\normalsize
\begin{tabular}{ccccc}
\hline
\textbf{Dataset} & \textbf{Tools} & \textbf{Correct} & \textbf{Incorrect} & \textbf{Accuracy} \\ \hline
\multirow{3}{*}{\begin{tabular}[c]{@{}c@{}}\sys\\ (200 CVEs)\end{tabular}} & \sys & 182 & 18 & 91\% \\
 & V0Finder & 66 & 134 & 33\% \\
 & VSZZ & 102 & 98 & 51\% \\ \hline
\multirow{2}{*}{\begin{tabular}[c]{@{}c@{}}SymBisect\\ (32 syzbot bugs)\end{tabular}} & \sys & 29 & 3 & 90.6\% \\
 & SymBisect & 24 & 8 & 75\% \\ \hline
\end{tabular}
}
}
\caption{\textbf{The results of BIC identification}} 
\label{evaluationresults_bisection}
\end{table}

\begin{table}[]
\centering
\scalebox{0.73}{
{\normalsize
\begin{tabular}{ccccccccc}
\hline
\textbf{Tools} & \textbf{TP} & \textbf{FP} & \textbf{FN}  & \textbf{Precision} & \textbf{Recall} & \textbf{F-1 Score} \\ \hline
~\sys     &4121     &151         &146            &96.5\%        &96.6\%        &96.5\%        \\
V0Finder & 1594  & 56  & 2748   & 96.6\% & 36.7\% & 53.2\% \\
VSZZ     & 4140 & 1660  & 85  & 71.4\% & 98.0\% & 82.6\% \\
\hline
\end{tabular}
}
}
\caption{\textbf{The results of vulnerable versions detection}}
\label{evaluationresults}
\end{table}

\begin{table}[]
\centering
\begin{tabular}{ccc}
\hline
\textbf{Year} & \textbf{\begin{tabular}[c]{@{}c@{}}Inaccurate\\ Cases\end{tabular}} & \textbf{\begin{tabular}[c]{@{}c@{}}Accuracy \end{tabular}} \\ \hline
2023 & 8 & 92\% \\
2024 & 10 & 90\% \\ \hline
\end{tabular}
\caption{\textbf{The accuracy with cases in different years}}
\label{table:year_difference}
\end{table}

\cut{
\begin{figure}[h]
  \centering
  \includegraphics[width=0.9\linewidth]{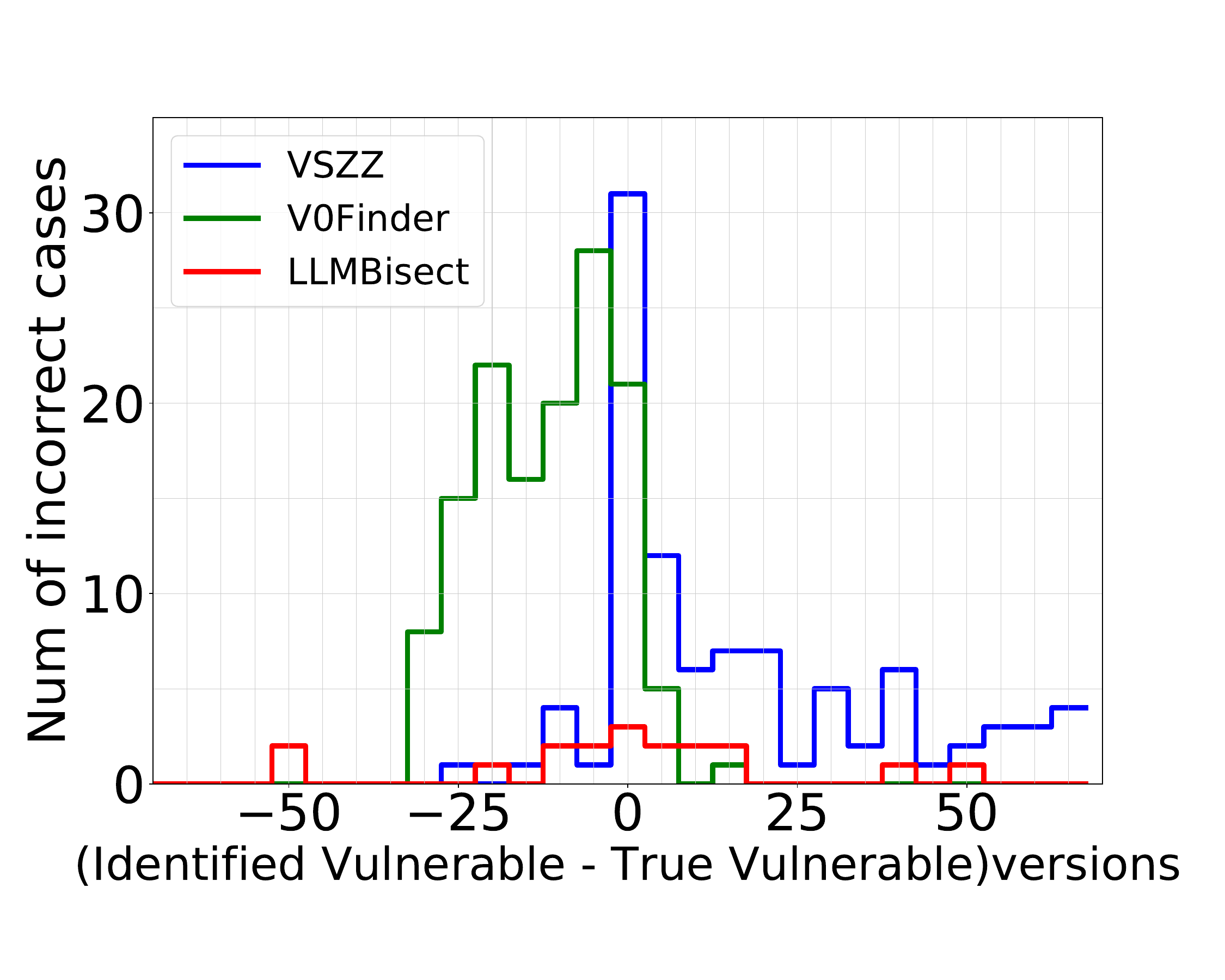}
  \vspace{-.16in}
  \caption{\textbf{Distribution of inaccurate cases over version distances}}
  \vspace{-.16in}
  \label{fig:distribution}
\end{figure}
}

\subsection{Accuracy of~\sys (RQ1)}
\label{sec:eval-acc}

\begin{table}[]
\scalebox{1}{
\begin{tabular}{ccc}
\hline
\textbf{Phase} & \textbf{Reason} & \textbf{Num} \\ \hline
\multirow{2}{*}{\begin{tabular}[c]{@{}c@{}}Candidate commit\\ Generation\end{tabular}} & BIC changed different files & 8 \\
 & Insufficient info in commit messages & 2 \\ \hline
BIC  Filtering & Not Pick groundtruth as final BIC & 4 \\ \hline
Result Finalization & Not Pick groundtruth as final BIC & 4 \\ \hline
\end{tabular}
}
\caption{\textbf{The reasons of \sys's inaccuracy}}
\label{reasonsoffailed}
\end{table}

\noindent\textbf{Accuracy of BIC Identification.}
Table~\ref{evaluationresults_bisection} shows the results of BIC identification with different tools,
\sys consistently achieves the highest accuracy of more than 90\%,
outperforming other state-of-the-art tools by significant margins (\ie 25.6\% - 58\%). Specifically, \sys accurately identified the correct BICs for the two motivating examples mentioned in Section~\ref{sec:motivation}.
Note that the comparison with SymBisect is based on SymBisect's own dataset due to its reliance on PoCs and specific vulnerability types, as mentioned previously in \S\ref{sec:eval}. 
These results show \sys's superior accuracy, even on dataset originally designed for other tools. We will describe the comparison results in detail in \S\ref{sec:eval-cmp}. Besides, Table \ref{table:year_difference} shows that there is nearly no difference in accuracy between cases from 2023 and those from 2024. This rules out the potential influence of the LLM’s pre-existing knowledge about the CVEs on the results, demonstrating the general applicability of our method.

\noindent\textbf{Inaccuracy Analysis.}
As shown in Table~\ref{evaluationresults_bisection}, \sys has 18 inaccurate bisection cases out of the 200 CVEs,
after inspecting each, we summarize 4 underlying reasons arising in 3 different phases of \sys (Fig.~\ref{fig:design}),
as listed in Table~\ref{reasonsoffailed}.
We now detail these reasons by phase.


\noindent\textit{Phase I: Candidate Generation.} 
\sys will miss the correct BIC (\ie false negative) if it is not included in the initial candidate list,
10 failure cases belong to this category.
Specifically, for 8 of them, the BIC and bug-fix commits modify completely different files,
making it difficult to recognize the correct BIC candidates without incurring a high cost
(\eg we need to enumerate virtually \emph{all} commits for all files in the codebase.).
In the remaining 2 cases,
the BIC and bug-fix commit modify different functions, structs, or variables within the same file,
however, our candidate generator fails to correlate them based on the bug-fix commit message,
which does not contain enough hints (\eg the vulnerable function name) to locate the remotely related BIC.

\noindent\textit{Phase II: BIC Filtering.}
In this phase, LLMs first try to identify (multiple) potential BICs from a specific generator's candidate list, then select \emph{one} BIC from multiple by comparing them.
We have 4 failure cases where the true BIC does not survive this filtering process.
Upon further investigation,
we found that the failure is mainly because of the excessive number of potential BICs to filter (\eg 84.25 on average for these 4 cases vs. 36.5 for all).
This confirms our design consideration (\S\ref{sec:design-motivation}) that more candidate commits can decrease the accuracy, besides increasing the costs.
We also observed that LLM's self-consistency issue contributes to 3 of these failure cases,
where the correct BIC can be selected in some LLM runs but not in others.

It is worth noting that we do not have any inaccurate cases in the pre-filtering phase (\eg LLMs fail to pick the correct BIC from the generator's candidate list at beginning), this confirms our observation (\S\ref{sec:design-motivation}) that LLM is less likely to make FNs when deciding whether an individual commit is a potential BIC.



\noindent\textit{Phase III: Result Finalization.}
Phase II selects one BIC from each of three generators' candidate lists, resulting in three final BIC candidates.
Then, the result finalizer further selects one BIC from these three.
5 failure cases are due to that the correct BIC does not survive this final ``1/3'' selection process.
We observed that the failure here is again related to LLMs' self-consistency (\eg correct BICs can survive in \emph{some} runs).

\subsection{Accuracy of Vulnerable Version Detection} 
One common application of bug bisection is to determine the software versions affected by a vulnerability, informing downstream developers for timely patch porting~\cite{zhang2021investigation}.
From this perspective, solely evaluating the accuracy of BIC identification has its limitations. For example, if a vulnerability is fixed in version 6.0 but introduced in version 5.0, a tool that identifies the introduction of the bug in version 5.1 or 5.19 would both be considered inaccurate from the perspective of BIC identification accuracy. However, the impact of such inaccuracies on downstream users can vary significantly.

\begin{figure}[h]
  \centering
  \includegraphics[width=0.95\linewidth]{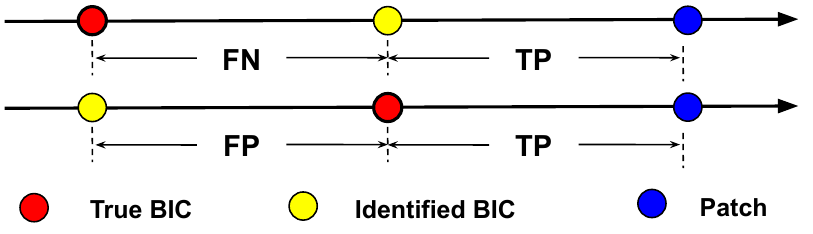}
  \caption{\textbf{Explanation of TP/FP/FN}}
  \label{fig:TP-FP-FN}
\end{figure}

Therefore, in addition to verifying whether our tool accurately identifies bug-inducing commits, we also evaluate the accuracy of identifying vulnerable versions. Specifically, once the BIC is determined, we can identify all vulnerable versions on the Linux mainline branch, i.e., versions between the BIC and the patch, considering only major releases such as v5.0, v5.1. By comparing the vulnerable versions derived from the true BIC with those derived from the BIC identified by our tool, we calculate the tool’s false positives (FP), false negatives (FN), and true positives (TP) for this task.  As Fig. \ref{fig:TP-FP-FN} shows, once we identify the BIC, we can determine the numbers of TP, FN, and FP based on its relative position to the true BIC and the patch commit. However, the number of TNs depends on the manually selected starting point (e.g., whether we start counting from v2.6 or v4.0) and is not a fixed value. Therefore, TNs are not included in our statistics.

\begin{figure}[h]
  \centering
  \includegraphics[width=0.9\linewidth]{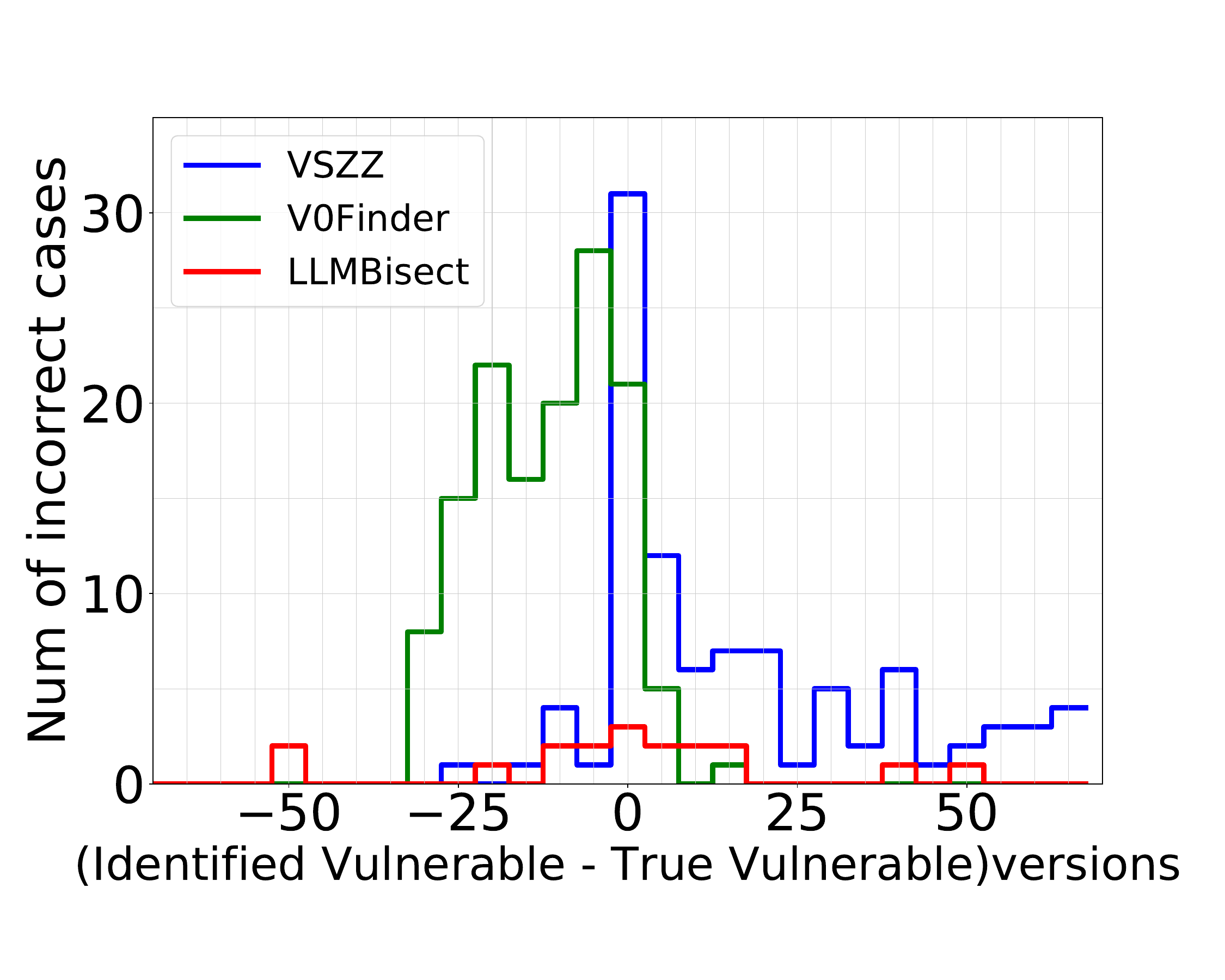}
  \vspace{-.16in}
  \caption{\textbf{Distribution of inaccurate cases over version distances}}
  \vspace{-.16in}
  \label{fig:distribution}
\end{figure}

As shown in Table \ref{evaluationresults},  ~\sys achieves an overall F-1 score of 96.5\%,  much higher than all existing tools. Note that this evaluation is performed on a per-bug-version-pair basis.

Fig. \ref{fig:distribution} shows the distribution of inaccurate cases for different methods in terms of FP/FN versions. The X-axis represents the number of FP or FN vulnerable versions for each case (e.g., 10 indicates a case where the method produced 10 false positive vulnerable versions, and -5 indicates a case where the method produced 5 false negative vulnerable versions). Note that, as shown in Fig.~\ref{fig:TP-FP-FN}, a single method cannot produce both FP and FN for the same case.

We group the inaccurate cases into intervals of 5 based on their FP/FN counts and plot the number of cases in each group on the Y-axis. From the figure, we can observe that VSZZ produces a large number of false positive versions, V0Finder generates many false negative versions, whereas \sys significantly reduces both false positives and false negatives.


\subsection{Comparison against SOTA Tools (RQ2)}
\label{sec:eval-cmp}

As shown in Table~\ref{evaluationresults_bisection} and Fig.~\ref{fig:distribution},   
\sys significantly outperforms other state-of-the-art tools regarding accuracy.
It achieves higher accuracy (91\% compared to the 41.5\% average of preceding tools in our dataset) in BIC identification and superior F1 scores (96.5\% as opposed to 67.9\%) compared to all prior tools. Remarkably, \sys even demonstrates much better performance than SymBisect on SymBisect’s own evaluation dataset.
In this section, we provide an in-depth analysis of these tools' inaccuracies and how \sys improves over them.

\noindent\textbf{V0Finder.}
V0Finder treats the pre-patched version of functions modified in the bug-fix commit as vulnerable,
it then compares it to all previous versions syntactically, by essentially a \emph{whole-function} strict string match with certain abstraction and normalization.
All identical historical versions will also be treated as vulnerable,
while the BIC is the commit turning a non-vulnerable version into vulnerable.
We detail V0Finder's weaknesses as follows.

\cut{
\begin{figure}[t]
\begin{center}  \includegraphics[width=1\columnwidth]{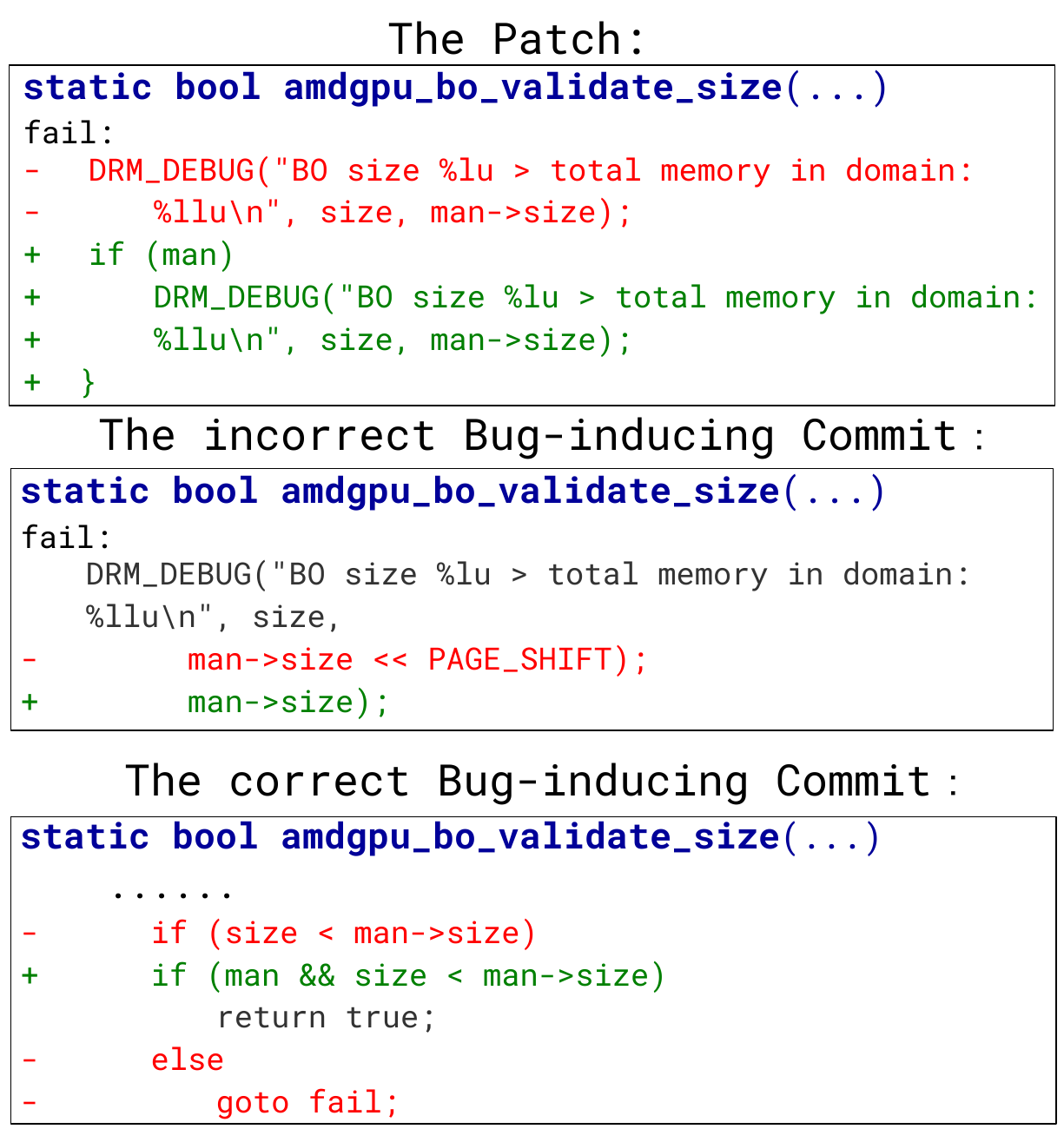}
\end{center}
\captionsetup{justification=centering}
\vspace{-.2in}
 \caption{Case study of VSZZ FP}
 \label{fig:vszz example}
 \vspace{-.1in}
\end{figure}
}


\noindent\textit{Flaw 1.}
Similar to \emph{Flaw 2} of VSZZ, the BIC may not make changes to the same functions as in the bug-fix commit (\eg 19 such cases in our dataset),
rendering V0Finder's patch-function-based BIC probing invalid. 


\noindent\textit{Flaw 2.}
V0Finder's syntactical similarity calculation is unaware of semantics and vulnerability logic.
Consequently, it will likely identify a historical commit as the BIC wrongly,
as long as it makes \emph{any} changes (that cannot be normalized or abstracted away by V0Finder's string matching algorithm) in the function patched by the bug fix,
These changes may not relate to the vulnerability at all (\eg not on the critical lines of the vulnerability) --- 84 of V0Finder's inaccurate cases are due to this,
or relate to but not introduce the vulnerability --- 31 failure cases are due to this.




As mentioned before,
\sys addresses these shortcomings by making decisions based on the understanding of the vulnerability logic with the help of LLMs
and its comprehensive consideration of the patch contexts. As a result, \sys resolves 117 out of 134 V0Finder's inaccurate cases.

\begin{table}[]
\resizebox{0.48\textwidth}{!}{
\begin{tabular}{ccc}
\hline
\textbf{Reason} & \textbf{\begin{tabular}[c]{@{}c@{}}Inaccurate\\ Cases\end{tabular}} & \textbf{\begin{tabular}[c]{@{}c@{}}Solved \\ in \sys\end{tabular}} \\ \hline
BIC changed different functions & 19 & 9 \\
Not identified critical lines & 84 & 80 \\
Flawed Heuristic & 31 & 28 \\ \hline
Total & 134 & 117 \\ \hline
\end{tabular}
}
\caption{\textbf{The reasons of V0Finder method failed}}
\label{reasonsofv0finderfailed}
\end{table}

\begin{table}[]
\resizebox{0.48\textwidth}{!}{
\begin{tabular}{ccc}
\hline
\textbf{Reason} & \textbf{\begin{tabular}[c]{@{}c@{}}Inaccurate\\ Cases\end{tabular}} & \textbf{\begin{tabular}[c]{@{}c@{}}Solved \\ in \sys\end{tabular}} \\ \hline
BIC changed different functions & 19 & 9 \\
Only focus on deleted lines & 28 & 27 \\
Not identified critical lines & 6 & 5 \\
Flawed Heuristic & 45 & 40 \\ \hline
Total & 98 & 81 \\ \hline
\end{tabular}
}
\caption{\textbf{The reasons of VSZZ method failed}}
\vspace{-.2in}
\label{reasonsofvszzfailed}
\end{table}

\begin{table}[]
\resizebox{0.48\textwidth}{!}{
\begin{tabular}{ccc}
\hline
\textbf{Reason} & \textbf{\begin{tabular}[c]{@{}c@{}}Inaccurate\\ Cases\end{tabular}} & \textbf{\begin{tabular}[c]{@{}c@{}}Solved \\ in \sys\end{tabular}} \\ \hline
Under-constrained Symbolization & 5 & 4 \\
Scalability & 3 & 3 \\
\hline
Total & 8 & 7 \\ \hline
\end{tabular}
}
\caption{\textbf{The reasons of SymBisect method failed}}
\vspace{-.2in}
\label{reasonsofsymbisectfailed}
\end{table}

\noindent\textbf{VSZZ.}
VSZZ identifies the BIC as the earliest commit that initializes the lines deleted by the bug-fix commit.
If the bug-fix does not delete any lines,
the commit initializing the file modified by the bug-fix will be treated as the BIC.
We group VSZZ's inaccurate cases based on flaws in this heuristic algorithm
and discuss how \sys addresses them.

\noindent\textit{Flaw 1.}
VSZZ fundamentally assumes that deleted lines in the bug-fix commit are related to the vulnerability's root cause,
so the BIC must introduce these lines.
However, the BIC can actually be within completely different functions (\eg \emph{19} such cases in our dataset) or irrelevant to those deleted lines (\emph{6} such cases in our dataset).

\noindent\textit{Flaw 2.}
The bug-fix commit can have no deleted lines, in this case, the heuristic of ``treating the line-initialization commit as BIC'' is oversimplified and highly inaccurate.
28 of bug-fix commits in our dataset have no deleted lines.

\noindent\textit{Flaw 3.}
The BIC may \emph{modify} but \emph{not initialize} the deleted lines in the bug-fix commit, 
violating VSZZ's heuristic. 
We observed \emph{45} such cases in our dataset.

The above flaws stem from VSZZ's reliance on hardcoded, simplified, and code-oriented heuristics.
\sys, on the other hand, utilizes LLM's deep and flexible understanding of vulnerability logic (\eg recognize critical lines) to identify BICs,
with minimal assumptions, \eg the presence (\emph{Flaw 2}) and significance (\emph{Flaw 1}) of deleted lines and BIC's operation (\emph{Flaw 3}.).
Furthermore, \sys takes advantage of full patch context, including the commit messages, to extract valuable information for BIC locating, significantly addressing \emph{Flaw 1}.
As a result, \sys resolves 81 out of 98 VSZZ's inaccurate cases.

\noindent\textbf{SymBisect.}
SymBisect decides whether a specific vulnerability affects a software version with under-constrained symbolic execution, guided by hints extracted from PoC execution traces for better scalability.
Despite its reliance on PoC and limited support for vulnerability types,
we identify issues impacting its accuracy on its own evaluation dataset (that we use for our comparison).

\noindent\textit{Flaw 1.}
Under-constrained symbolic execution assumes overly relaxed constraints (and often infeasible) of program variables unknown in its analysis scope, \eg global variables initialized outside of the local analyzed function(s).
This results in over-approximation of program behaviors, for instance, a software version can wrongly be recognized as vulnerable.
SymBisect fails in 5 cases in our dataset due to this reason.


\noindent\textit{Flaw 2.}
Symbolic execution is known to be expensive.
To address the scalability issue,
SymBisect utilizes information (\eg promising paths) extracted from PoC execution traces to guide its symbolic execution.
However, this guide may be incomplete or inaccurate,
leading to missed vulnerable paths and/or conditions, eventually causing inaccuracies in BIC identification.
We observed 3 such inaccurate cases in the SymBisect evaluation dataset.


\sys, unlike SymBisect, does not rely on the expensive symbolic execution for BIC identification.
Instead, its decision is based on LLM's profound understanding of the vulnerability logic, from both code changes and commit messages, avoiding the above difficulties.


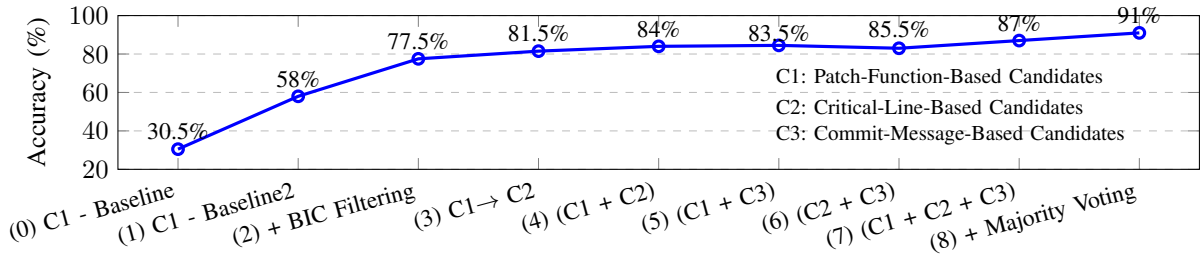
\begin{figure*}[ht]
    \centering


\begin{tikzpicture}[scale=1.0]
\begin{axis}[
    xlabel = {},
    ylabel={Accuracy (\%)},
    ymin=20, ymax=100,
    xmin=0.5, xmax=9.5,
    xtick={1,2,3,4,5,6,7,8,9},
    xticklabel style={
        font=\small,        
        rotate=15,          
        anchor=north east   
    },
    xticklabels={
      {(0) C1 - Baseline}, 
      {(1) C1 - Baseline2},
      {(2) + BIC Filtering},
      {(3) C1$\rightarrow$ C2},
      {(4) (C1 + C2)},
      {(5) (C1 + C3)},
      {(6) (C2 + C3)},
      {(7) (C1 + C2 + C3)},
      {(8) + Majority Voting}
    },
    nodes near coords,      
    nodes near coords align={vertical}, 
    every node near coord/.append style={
        font=\small,        
    },
    point meta=explicit symbolic, 
    ymajorgrids=true,
    grid style=dashed,
    width=0.88\textwidth,  
    height=0.2\textwidth
]
\addplot[
    color=blue,
    mark=o,
    line width=1.2pt
] 
coordinates {
    (1,30.5)
    (2,58)
    (3,77.5)
    (4,81.5)
    (5,84)
    (6,84.5)
    (7,83) 
    (8,87)
    (9,91)
};

\addplot[
    only marks,
    mark=none,
    point meta=explicit symbolic
] 
coordinates {
    (1,30.5) [30.5\%]
    (2,58) [58\%]
    (3,77.5) [77.5\%]
    (4,81.5) [81.5\%]
    (5,84)   [84\%]
    (6,81.5) [83.5\%]
    (7,84)   [85.5\%]
    (8,87)   [87\%]
    (9,91)   [91\%]
};
\node[anchor=south west] at (rel axis cs: 0.6, 0.50) {\footnotesize C1: Patch-Function-Based Candidates};
\node[anchor=south west] at (rel axis cs: 0.6, 0.30) {\footnotesize C2: Critical-Line-Based Candidates};
\node[anchor=south west] at (rel axis cs: 0.6, 0.10) {\footnotesize C3: Commit-Message-Based Candidates};
\end{axis}
\end{tikzpicture}

    \vspace{-.1in}
    \caption{\textbf{Ablation Study with Different Design Points}}
    \label{fig:ablation}
\end{figure*}

\subsection{Ablation Study (RQ3)}
\label{sec:eval-ablation}

\subsubsection{Effectiveness of Design Points}

As discussed in \S\ref{sec:design-motivation},
our final design results from multiple iterations and refinements of a baseline workflow.
During this process, we adopt different effective design points that \emph{all} improve \sys's accuracy.
To demonstrate it, we start with the baseline method and gradually integrate each of our design points,
observing the change in BIC detection accuracy.
We show the results in Fig.~\ref{fig:ablation}, as can be seen, the accuracy steadily improves as more design points are adopted (\eg from 30.5\% to 91\%).
In the remainder of this section,
we detail the reasons behind these improvements by analyzing each intermediate configuration in Fig.~\ref{fig:ablation}.



\noindent\textit{(0) The Baseline Method.}
As described in \S\ref{sec:design-motivation},
the most straightforward baseline method inspired by existing work is to let LLM inspect each commit (reverse chronologically) that touches the same function(s), i.e., candidates are generated using the patch-function-based generator alone.
The first identified BIC will be output as the final result.
As analyzed in \S\ref{sec:design-motivation},
this approach has a low accuracy (30.5\% in Fig.~\ref{fig:ablation}) mainly due to LLM's high false positive rate in single-commit BIC decision and missing true BIC with single generator.

\noindent\textit{(1) A Second Baseline Method.} We also consider an alternative baseline method where we pick BIC candidates from the N most recent relevant commits that modify the patched function(s), where N is chosen such that we do not overfill the context window of the LLM. We will have the LLM pick a single commit from them as the BIC. 
While a plausible design, we note that it has two major conceptual limitations:
1. The most recent N relevant patches may not include the true BIC if it resides early in the long commit history.
2. A long input (near the context window limit) is known to degrade LLM's performance \cite{hsieh2024ruler,li2404long,liu2023lost}. It is especially challenging given the many candidate BIC commits that are interconnected (modifying the same function). 
As shown in Fig.~\ref{fig:ablation}, this method (Baseline2) achieved an overall accuracy of 58\% (116/200), outperforming the previous baseline but still falling short of our proposed approach. In 51 cases, the true BIC was absent from the LLM's candidate set, either because it was not identified as relevant or because it was excluded due to the context window limit (the first limitation noted above). In another 33 cases, the true BIC was present among the candidates, but the LLM selected incorrectly (the second limitation).




\noindent\textit{(2) Added: BIC Filtering.}
We then adopt the BIC comparative filtering process (\S\ref{sec:bic-selector}),
where all potential BICs are identified and then compared by the LLM to determine the most likely one.
As shown in Fig.~\ref{fig:ablation}, this significantly improve the accuracy compared to the strawman workflow (30.5\% $\rightarrow$ 77.5\%).




\noindent\textit{(3) Replaced: C1 $\rightarrow$ C2.}
Patch-function-based candidate generation (\ie \emph{C1} in Fig.~\ref{fig:ablation}) can result in too many candidates,
confusing the LLM and eventually reducing accuracy.
We show that a more fine-grained critical-line-based strategy (\emph{C2} in Fig.~\ref{fig:ablation} to replace \emph{C1}, detailed in \S\ref{sec:candidate-generation}) increases the accuracy from 77.5\% to 81.5\%.



\noindent\textit{(4) Added: Result Finalizer.}
As discussed in \S\ref{sec:design-motivation},
critical-line-based candidate generation (\emph{C2} in Fig.~\ref{fig:ablation}) is more precise,
however, it can also miss true BICs if some critical lines are missed.
Our solution is to combine \emph{C2} and \emph{C1} with the \emph{result finalizer} (\S\ref{sec:result-merge}),
this design further improves the accuracy (81.5\% $\rightarrow$ 84\% in Fig.~\ref{fig:ablation}).

\noindent\textit{(5) Replaced: C1+C2 $\rightarrow$ C1+C3.}
We also tried C1+C3 and got similar results (83.5\%) compared to C1+C2.

\noindent\textit{(6) Replaced: C1+C2 $\rightarrow$ C2+C3.}
C2+C3 and got similar results (85.5\%) compared to C1+C2.


\noindent\textit{(7) Added: Commit-Message-Based Candidate Generation.}
Neither \emph{C1} nor \emph{C2} captures BICs having no code overlaps with the corresponding bug-fix commits,
as mentioned in \S\ref{sec:design-motivation}.
We thus develop another strategy that seeks implicitly connected BICs from the commit messages of the bug-fix (\emph{C3} in Fig.~\ref{fig:ablation}, detailed in \S\ref{sec:candidate-generation}).
As shown in Fig.~\ref{fig:ablation},
this improves the accuracy to 87\% from the previous configuration.




\noindent\textit{8) Added: Majority Voting.}
As mentioned before (\S\ref{sec:eval-acc}),
the well-known self-consistency issue of LLMs can negatively impact our accuracy,
when the correct decision is not yielded in the first run.
To address this,
we incorporate the \emph{majority voting} mechanism which selects the most frequent answer among multiple LLM runs in the result finalizer.
This further improves \sys's accuracy compared to the previous configuration (\ie 87\% $\rightarrow$ 91\% in Fig.~\ref{fig:ablation}).

Importantly, our approach is not a simple application of LLMs. As shown above, an intuitive LLM-based method yields low accuracy (30.5\%). We identified the shortcomings of such baseline approaches (outlined as four challenges) and addressed them through a structured workflow. Some design components, such as the three complementary candidate commit generators, are not LLM-specific (though one uses an LLM). We carefully assigned tasks to LLMs only where they are most effective. The ablation study confirms the advantages of this design.




\begin{figure}[h]
  \centering
  \includegraphics[width=1\linewidth]{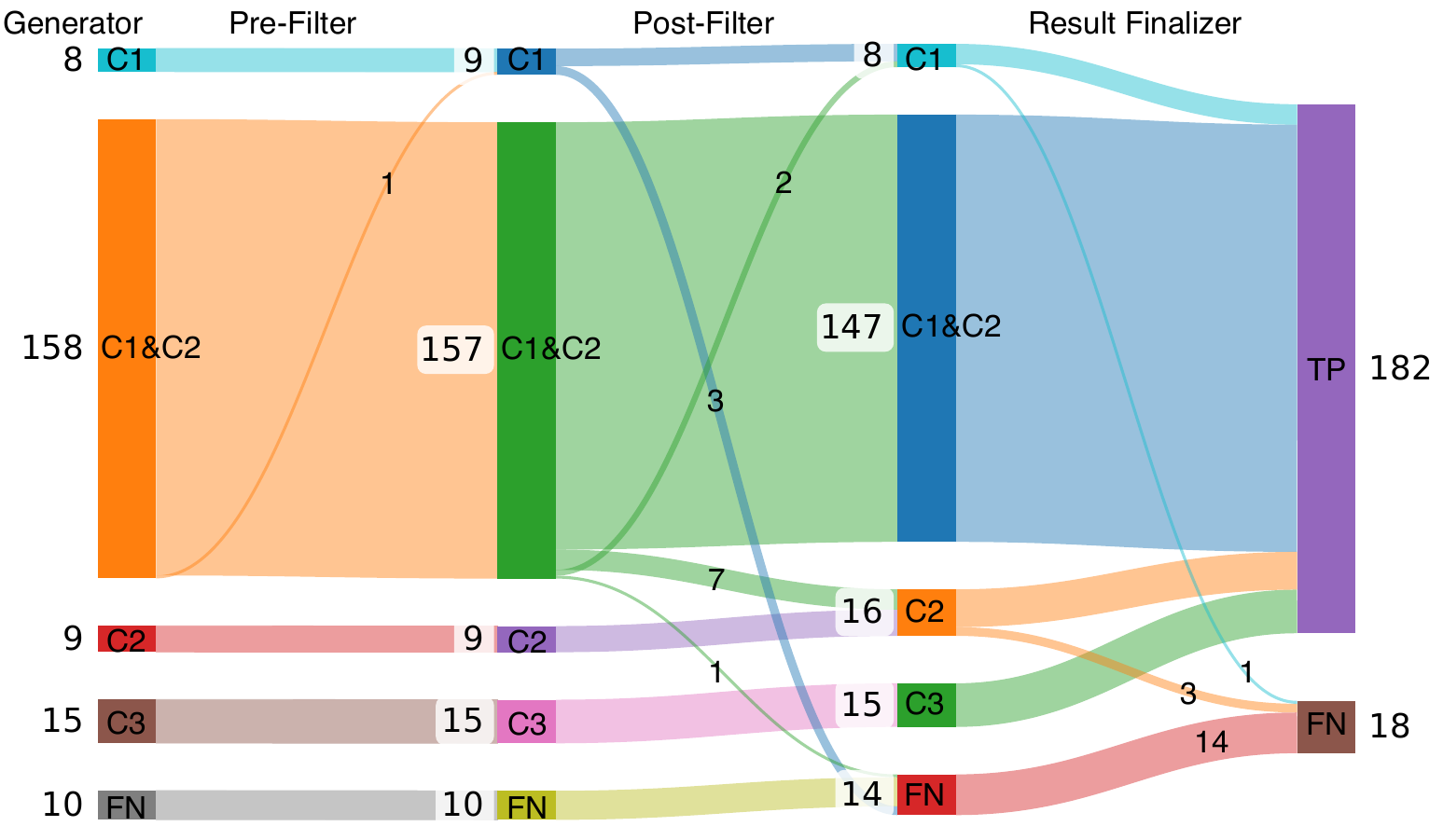}
  \caption{\textbf{Flow of Ground-Truth Bug-Inducing Commits}
  }
  \label{fig:TP_Flow}
\end{figure}

\begin{figure}[h]
  \centering
  \includegraphics[width=\linewidth]{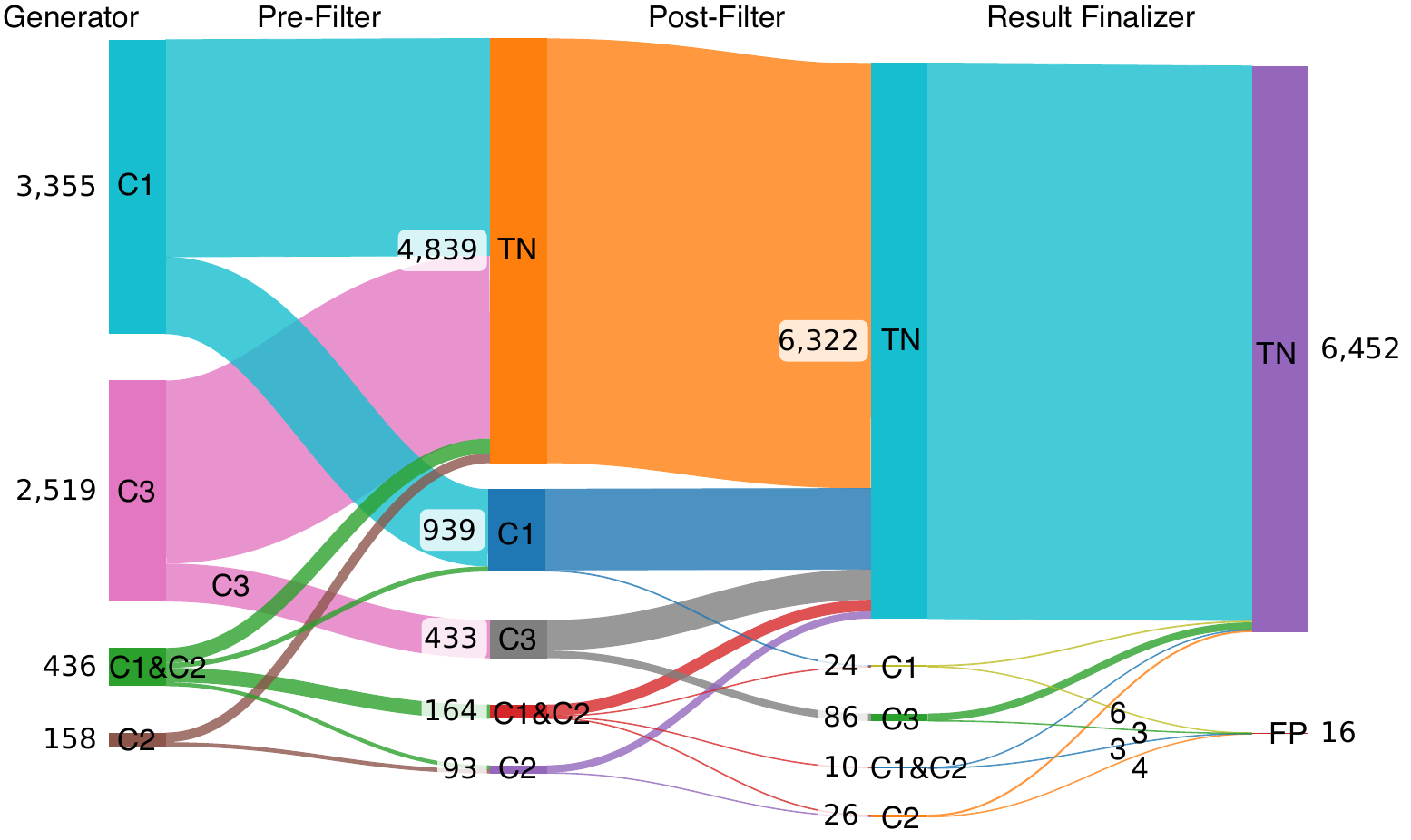}
  \caption{\textbf{Flow of False Bug-Inducing Commits}
  }
  \label{fig:FP_Flow}
\end{figure}

\subsubsection{Breakdown by Components and Phases}
To more effectively illustrate the relationships between different candidate generators and understand their individual importance, we present two Sankey diagrams in Fig.~\ref{fig:TP_Flow} and Fig.~\ref{fig:FP_Flow}.
Fig.~\ref{fig:TP_Flow} visualizes recall losses, showing how the correct BICs are wrongly filtered out or overlooked (\ie false negatives) by different methods at each phase.
In contrast, Fig.~\ref{fig:FP_Flow} tracks how incorrect BICs (\ie false positives) are progressively pruned across multiple pipeline stages.
As previously termed in Fig.~\ref{fig:ablation}, C1, C2 and C3 refer to the candidates produced by our three generators (\ie function-, critical-line-, and commit-message-based), respectively.
The label ``C1 \& C2'' indicates the \emph{intersection} of C1 and C2 (\eg in Fig.~\ref{fig:TP_Flow}, 158 true BICs were initially identified by both the function- and critical-line-based generators).
Note that C3 by design has no overlaps with either C1 or C2,
as C3 intends to identify commits that are otherwise missed by C1 and C2 (see Challenge \#2 in the \S\ref{sec:design-motivation}).
It is worth mentioning that each bug has exactly one true BIC per our definition.
Thus, TPs and FNs add up to 200 (\ie total number of our evaluated bugs) in Fig.~\ref{fig:TP_Flow}.
However, since each generator may initially identify multiple candidate BICs for a given bug,
FPs in the early stages significantly exceed 200 in Fig.~\ref{fig:FP_Flow}.
Note that the final numbers of FN (in Fig.~\ref{fig:TP_Flow}) and FP (in Fig.~\ref{fig:FP_Flow}) are not equal, i.e., 18 vs. 16. 
This is because in two bugs, \sys failed to produce any result (none of the candidate commits were identified as the BIC). These cases are counted as FNs but not as FPs.

We have the following observations:

\noindent\textbullet\ C1, C2, and C3 effectively \textbf{complement each other} by covering the true BICs that the others miss, collectively contributing to the high overall accuracy of our approach (Observation \#4 in Design Section).
This is shown in Fig.~\ref{fig:TP_Flow}. Using only C1, the accuracy is 77.5\% (155 out of 200), and using only C2, the accuracy is 81.5\% (163 out of 200).
For instance, 15 true BICs are \emph{exclusively} identified by only C3 in the very first phase and retained throughout the remainder of the pipeline,
contributing to a 7.5\% improvement in overall accuracy.
Without C3, we would have 25 instead of currently 10 missed true BICs in Phase I.
This result confirms Observation \#2 and highlights the effectiveness of Solution \#2 described in \S\ref{sec:design-motivation}.
Similarly, both C1 and C2 have their own indispensable contributions across the pipeline, supporting the design rationale behind Solution \#1.


\noindent\textbullet\ Strong True BIC Retention.
Our pipeline design demonstrates a strong ability to retain the true BICs at every stage, achieving a high end-to-end TP rate.
As shown in Fig.~\ref{fig:TP_Flow}, the combined filtering across all three stages,
\ie Pre-Filter, Post-Filter, and Result Finalizer,
mistakenly discarded only 8 true BICs,
yielding a high recall of 95.8\% (182/190).
Particularly noteworthy is the performance of the LLM-based method in the Pre-Filter stage: 
while eliminating 4,839 FP candidates (Fig.~\ref{fig:FP_Flow}),
it did not discard a single correct BIC initially identified by generators.
Such a high recall (or low false negative) is likely due to its capacity to align high-level natural language intent (from commit messages) with low-level code modifications, a capability vital for tracing causal relationships between BICs and patches.
In addition, the Post-Filter and Result Finalizer stages each missed only 4 true BICs, highlighting the powerful comparative reasoning ability of LLMs and providing strong empirical support for Observation \#3.

\noindent\textbullet\ Effective False Positive Filtering. Fig.~\ref{fig:FP_Flow} shows that there are many incorrect BICs (\ie false positives) generated at the beginning. 
Specifically, the total number of FP candidates initially exceeds 6,400 across all three strategies, and the majority of FPs are contributed by C1, consistent with our Challenge/Observation/Solution \#1. 
More than 1,600 false positives remain after Pre-Filtering --- a clear reflection of Challenge \#3. It is likely due to the lack of a well-defined notion of ``bug-inducing commits'' to LLMs --- any commit that appears related to the vulnerability is considered bug-inducing. 
Nevertheless, our design effectively filtered out false positives (FPs) throughout the pipeline.
In the Pre-Filter stage, 75\% of initial FPs (\ie 4,839 candidates) are successfully filtered out.
The Post-Filter stage additionally removes 1,483 FPs, and eventually,
only 16 FP candidates remain after the Result Finalizer stage. 
We can see the contribution of false positives from C1, C2, C3 is as follows:  
C1-only: 6,  C2-only: 4,  C1\&C2: 3,  C3-only: 3.
This result again confirms LLM's powerful comparative reasoning (Observation \#3),
which we effectively leverage in our pipeline design (Solution \#3).)

\noindent\textbullet\ Internal consistency.
Fig.~\ref{fig:TP_Flow} illustrates that in 147 cases, both C1 and C2 include the correct top candidate, and in all such cases, the Result Finalizer selects them correctly. Additionally, there are 8 cases where only C1 contains the correct BIC, 16 cases where only C2 does, and 15 cases where only C3 does. Overall, we observe that the correct top candidate is selected in the vast majority of cases. The only exceptions are 1 case where the top candidate appears only in C1, and 3 cases where it appears only in C2 --- these candidates are correct BICs, but the Result Finalizer does not select them.

\begin{table}[]
\resizebox{0.48\textwidth}{!}{
\begin{tabular}{ccc}
\hline
\textbf{\begin{tabular}[c]{@{}c@{}}Patch\\ Information\end{tabular}} & \textbf{\begin{tabular}[c]{@{}c@{}}Inaccurate\\ Cases\end{tabular}} & \textbf{\begin{tabular}[c]{@{}c@{}}Accuracy \end{tabular}} \\ \hline
Commit Message+ Code Change & 19 & 91.0\% \\
Code Change & 58 & 71.0\% \\ \hline
\end{tabular}
}
\caption{\textbf{The accuracy with/without commit message}}
\vspace{-.2in}
\label{table:ab-message}
\end{table}

\subsubsection{The Role of Commit Messages}
\label{sec:commit_messages}

One of \sys's major advantages is its utilization of the full patch information, including both code changes and natural language commit messages.
Besides \emph{C3} in Fig.~\ref{fig:ablation} for candidate initialization,
commit messages also help LLMs make more informed decisions when inspecting each commit for BIC identification.
To quantitatively understand the commit message's impact,
we strip the commit messages of all commits and re-run our evaluation.
Note that the impact is multi-front: 
(1) the commit-message-based generator basically no longer works,
(2) the critical-line-based generator is substantially weaker because the LLM can no longer benefit from the commit messages to understand the logic of the bug,
and (3) the selection of the BICs is also weaker because the LLM can no longer benefit from the description of the purpose of the candidate commits.
As shown in Table~\ref{table:ab-message},
the gap in accuracy is significant: 71\% vs 91\%.

\begin{table*}[]
\centering
\begin{tabular}{|c|cccccc|cccc|c|c|}
\hline
Candidates Size & 3 & 3 & 3 & 3 & 3 & 3 & 2 & 2 & 2 & 2 & 1 & 0 \\ \hline
Count of each BIC & (7, 0, 0) & (6,1,0) & (5,2,0) & (4,3,0) & (3,2,2) & (4,2,1) & (7, 0) & (6,1) & (5,2) & (4,3) & (7) & () \\ \hline
Num of Cases & 76 & 9 & 6 & 2 & 1 & 1 & 92 & 2 & 1 & 1 & 7 & 2 \\ \hline
\end{tabular}
\caption{\textbf{Majority voting inter-consistency}}
\vspace{-.1in}
\label{majorityvoting}
\end{table*}

\subsubsection{Consistency in Result Finalizer's majority voting}

Table \ref{majorityvoting} presents the statistics of the majority voting after repeating the queries seven times.
Note that Result Finalizer receives up to three candidate inputs, each from a different candidate generator. For 95 cases where we had three candidates, the LLM produced a vastly consistent 7:0:0 vote in 80\% (76 cases). An additional 9 cases resulted in a 6:1 vote.
When the number of candidates is two, the LLM produced an overwhelming 7:0 vote in 95.8\% (92/96) of the cases. As the number of candidates decreases, the stability of the LLM's output increases --- an observation consistent with what we discussed in Challenge \#1.

\begin{figure}[h]
  \centering
  \includegraphics[width=0.8\linewidth]{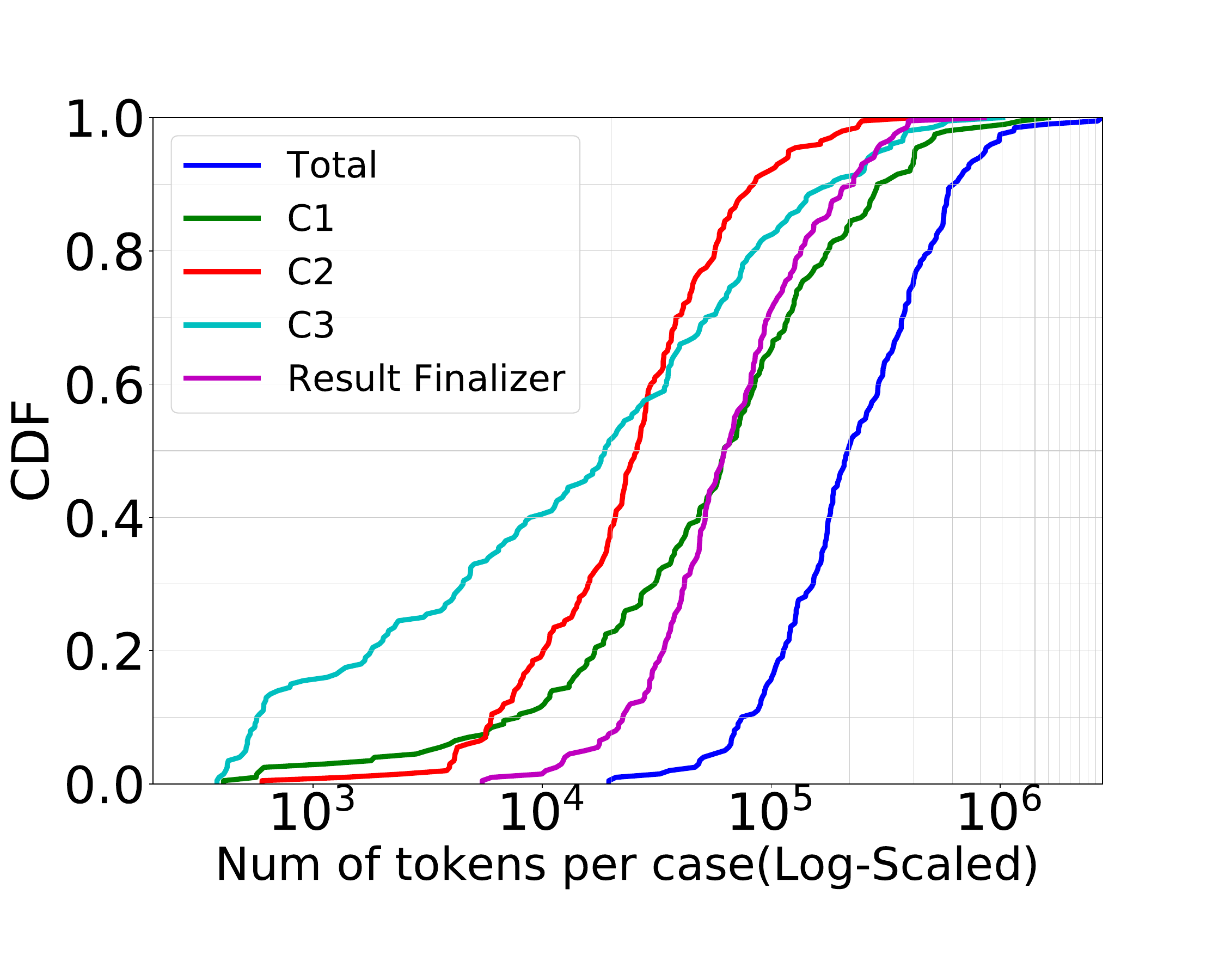}
  \vspace{-.2in}
  \caption{\textbf{Distribution of token cost per case}}
  \vspace{-.2in}
  \label{fig:distribution_token}
\end{figure}

\begin{table}[]
\centering
\begin{tabular}{ccc}
\hline
\textbf{Model} & \textbf{\begin{tabular}[c]{@{}c@{}}Inaccurate\\ Cases\end{tabular}} & \textbf{\begin{tabular}[c]{@{}c@{}}Accuracy \end{tabular}} \\ \hline
OpenAI o1 & 19 & 91.0\% \\
GPT-4o & 69 & 65.5\% \\
LLama3.2 & 58 & 71.0\% \\ \hline
\end{tabular}
\caption{\textbf{The accuracy with different LLM models}}
\vspace{-.2in}
\label{table:ab-model}
\end{table}

\subsection{Different LLMs}
\sys's design is agnostic to the underlying LLM; Nonetheless, we conduct a comparative evaluation by swapping between three widely used LLMs: OpenAI o1, GPT-4o, and LLama 3, covering both commercial and open-source models.
The evaluation results are summarized in Table~\ref{table:ab-model}.
As can be seen, OpenAI o1 achieves the highest accuracy (91\%) likely due to its enhanced reasoning capability, followed by LLama 3 (71\%) and GPT-4o (65.5\%). 
We found that the majority of inaccuracies occur during the process of comparing multiple suspected BICs and selecting the final result (specifically, during the Post-Filtering and Result Finalizer stages). For example, GPT-4o produced a total of 49 inaccurate cases across these two steps. This suggests a gap for different models in tasks which requires extensive reasoning on multiple code snippets and commit message.

\subsection{Token cost (RQ4)}
\label{sec:token_cost}
For each case using the O1 model, the minimum token usage was 19,557 tokens (approximately \$0.30 at the time of writing), and the maximum was 2,772,509 tokens (approximately \$41). The median was 219,523 tokens (approximately \$3.3 with the o1 model), and the average was 330,426 tokens (approximately \$4.9). Overall, this cost is acceptable compared to the expense of hiring professionals for manual analysis and the potential risks posed by N-day vulnerabilities. 

We further break down the cost by methods, which is shown in Fig. \ref{fig:distribution_token}. 
As expected, the patch-function-based method (that generates C1) consumes significantly more tokens than the critical-line-based method and the commit-message-based method.
The Result Finalizer stage also consumes a significant number of tokens because of the majority voting that repeats the experiment seven times.

\subsection{Case Study.}

We revisit the example in Fig.\ref{fig:motivating example} to illustrate how \sys identifies the correct bug-inducing commit (BIC), highlighting key features, the strengths and limitations of LLMs, and insights into their behavior.

Initially, generators C1, C2, and C3 produce 3, 2, and 35 BIC candidates, respectively. Only C3 includes the correct BIC, as it appears in the commit message but modifies different functions/structs than the patch, making it inaccessible to C1 and C2.

\PP{High False Positives}
After the Pre-Filter step, 1, 1, and 20 candidates remain from C1, C2, and C3, respectively. The LLM flags all 22 as potential BICs, reflecting its tendency toward high false positives (Design Challenge \#3).
For example, commit c19ffe00fed6 is incorrectly identified as a candidate BIC because it significantly refactored the patched function \texttt{gsm\_modem\_update()} (see Appendix Fig.\ref{fig:motivating example FP}).

\PP{Comparative Reasoning Ability}
In the Post-Filter stage, \sys selects the most likely BIC among candidates. The LLM correctly picks the true BIC from the 20 candidates in C3, demonstrating strong comparative reasoning. Rather than making isolated binary decisions, it evaluates relative differences, showing a deeper understanding of vulnerability logic in context.

In the Result Finalizer stage, \sys selects the correct BIC from the three final candidates using majority voting --- all seven runs agree. The LLM’s explanation aligns closely with the actual root cause:

\textit{``The patch description clearly indicates that the crash occurs because data can still be queued and the 'kick\_timer' restarted after gsm\_cleanup\_mux() has already begun tearing down resources. ......
c568f7086c6e, is where the problematic timer-based re-queuing mechanism is added, but does not check if \texttt{gsm->dead} is set. This allowed updates (and modem status line changes) to be triggered after teardown, hence causing the race condition that the final patch fixes.''}

\PP{Importance of Both Commit Messages and Code Changes.}
A core strength of \sys is its ability to integrate information from both commit messages and code. Following the ablation study on the importance of commit messages, we find that removing either commit messages or code changes as input will lead the LLM to misidentify c19ffe00fed6 as the BIC in all seven runs. Below is an example LLM response that incorrectly summarizes the behavior of the true BIC.

\textit{``c568f7086c6e focuses on adding a kick\_timer to handle data transmission logic and does not appear to introduce the risk of calling gsm\_modem\_update after the gsm mux is marked dead.''}

\section{Limitations and Discussion}

\noindent\textbf{Incomplete BIC candidate generation.}
As reported in Table~\ref{reasonsoffailed}, there are 10 failure cases where our solution simply failed in the candidate generation phase. 
For 8 of them, the file containing the true BIC is not explicitly mentioned in the patch. Our current design is unable to handle such cases.
One potential strategy to solve this is to expand the scope of code context based on a dependency analysis, e.g., slicing, to identify more relevant functions beyond the patched ones. One can also extract less precise hints in the commit message (e.g., mentioning a module name) to constrain the search space. Retrieval-Augmented Generation (RAG) is one potential solution to extract additional relevant code context automatically.
We leave addressing these corner cases as future work.


\noindent\textbf{Non-determinism.}
As with most LLM-based solutions, the results are inherently non-deterministic due to sampling during generation. To mitigate this, common prompt strategies such as majority voting are often used, and we adopt this strategy in our work as well.
In cases where the LLM exposes confidence scores or output probabilities, an alternative is to select the response with the highest likelihood, though this depends on the interfaces provided by specific models.

\noindent\textbf{Dependency on the quality of commit messages.}
As we show in \S\ref{sec:commit_messages}, commit messages indeed provide significant benefits to the overall accuracy of the solution. On the flip side, it also means that our solution depends on the quality of the commit messages. As a result, we expect to see degraded performance when our solution is applied to projects where the commit messages are not as informative as those in the Linux kernel. Some possible mitigations are (1) perform better prompt engineering to extract more description about the bug before conducting bug bisection, and (2) leverage RAG to obtain more information about the patches and the bugs, e.g., from mailing lists or other sources.

\noindent\textbf{Advanced LLM Post-training \& Prompt Techniques.}
Several proven techniques can further boost the performance and cost-efficiency of \sys.  Fine-tuning and   \textit{Reinforcement learning from human feedback} (RLHF) can align responses to our intent ~\cite{ouyang2022instructgpt}. As mentioned above, \textit{Retrieval-Augmented Generation} (RAG)~\cite{lewis2020rag} can fetch the most relevant context (e.g., functions, relevant documents, mailing list threads) to provide additional code context or bug-related information, thereby improving performance.  
Advanced prompt and context design, such as Chain-of-Thought~\cite{wei2022chainofthought} and
\textit{Multi-agent LLM architectures}~\cite{superannotate2025llmagents} improve LLM in 
reasoning without extra training. These techniques could help the migration of \sys to weaker models and save costs.  We leave the exploration of these directions to future work.

\section{Related Work}

\noindent\textbf{The Application of LLMs in Program Analysis.} 
Recent research has explored the integration of LLMs into static analysis to enhance its effectiveness in code comprehension and bug findings \cite{wang_sanitizing_2024, wadhwa_core_2024, li2024enhancing, wang_llmsa_2024}. 
LLMs have also been employed to understand and generate code comments, documentation, and system logs, improving code readability and maintainability \cite{jiang_lilac_2024, li_only_2024, geng_large_2024}. The integration of LLMs in program analysis represents a significant advancement in software engineering, offering tools that enhance productivity, code quality, and security.

\noindent\textbf{PoC-based vulnerable version identification.} 
SymBisect~\cite{zhang2024symbisect} leverages under-constrained symbolic execution to determine whether a specific software version contains a given vulnerability, enabling the identification of BICs. 
However, SymBisect supports only specific types of functions and requires an existing PoC.
Dai et al.\cite{dai2021facilitating} proposed a PoC migration approach that takes an initial PoC as input and adapts it to identify other affected versions; however, it is specifically designed for user-space programs.

\noindent\textbf{SZZ Methods.} SZZ (short for Śliwerski, Zimmermann, and Zeller)~\cite{SZZ} is an algorithm designed to identify bug-inducing commits in version control systems, also called B-SZZ. It identifies earlier changes at the location of a bug fix as bug-inducing commits. However, its straightforward approach struggles to handle complex bugs effectively. To address this limitation, AG-SZZ~\cite{AG-SZZ} incorporates an annotation graph to exclude non-semantic changes, such as whitespace, comments, and formatting adjustments, thereby reducing false positives. MA-SZZ~\cite{MA-SZZ} further improves on this by filtering out meta-changes like branch modifications and file attribute updates, ensuring that only source code changes are analyzed. V-SZZ~\cite{vszz} expands the algorithm’s scope by targeting vulnerabilities introduced in earlier software versions. NEURAL-SZZ~\cite{n-szz} leverages a Heterogeneous Graph Attention Network (HAN) to capture semantic relationships between lines of code, enhancing precision in tracing bug origins. However, it is limited to Java and exhibits a relatively high false positive rate. Combining advanced techniques like NEURAL-SZZ and V-SZZ can significantly improve bug-tracing accuracy, while AG-SZZ and MA-SZZ remain practical solutions for simpler scenarios.

\noindent\textbf{Vulnerable code clone detection.}
Vulnerable code clone detection is a specialized type of code clone detection \cite{ain2019systematic,roy2009comparison,sajnani2016sourcerercc,shobha2021code,fang2020functional}. It involves identifying pieces of source code in software systems that are similar to or identical to code fragments known to have security vulnerabilities.
They usually perform similarity comparisons on what they define as vulnerability-related code (usually a few lines within the patch function or the entire function) \cite{VUDDY,jang2012redebug,bowman2020vgraph,xiao2020mvp,zou2017scvd,bowman2020vgraph,V0finder}.
However, rule-based code extraction and similarity-based solutions often fail to identify vulnerability-relevant code or confirm the presence of vulnerabilities, as they lack vulnerability comparison based on logical structures. Our evaluation demonstrates that such methods perform poorly on complex programs such as the Linux kernel.


\cut{
\noindent\textbf{Bug-report-based bisection.}
Locus \cite{wen2016locus} was the first method to identify BIC by leveraging token similarities extracted from bug reports \cite{wu2018changelocator,bhagwan2018orca}. 
Bug2Commit \cite{murali2021industry} combines features from bug reports and commits by averaging their vector representations. 
FONTE \cite{an2023fonte} detects BIC using test coverage. 
The state-of-the-art method, FONTE, achieves an accuracy rate of only 36\%.}
\section{Conclusion}

In conclusion, we introduced \sys,  a novel, LLM-driven bug bisection pipeline that effectively pinpoints bug-inducing commits in Linux kernel. By combining both code changes and commit-message insights, \sys overcomes the limitations of traditional patch-based methods, which often fail to capture the true scope and context of a vulnerability. 
Our results underscore the potential of large language models to streamline vulnerability detection, reducing the window in which attacks can occur.

\section{Ethics Considerations }


This research was conducted in alignment with recognized ethical guidelines, ensuring responsible practices in methodology, data handling, and reporting. Our work aims to enhance bug bisection and vulnerability identification in open-source software, ultimately helping developers and maintainers address security threats more effectively.

Our research is based on N-day vulnerabilities—specifically, bugs that have already been patched in the Linux mainline. Furthermore, we do not anticipate any adverse impact on individuals or groups, as our analysis is strictly limited to publicly available codebases and does not involve any personal or sensitive data.

In developing \sys, we utilized both closed-source and open-source large language models without incorporating any copyrighted or sensitive material.

\section*{Acknowledgment}

We thank the anonymous reviewers for their insightful comments and
valuable suggestions. This material is based upon work supported by
the National Science Foundation under Grant No. \#2155213 \& \#2247881
and the Defense Advanced Research Projects Agency (DARPA) under Agreement No. HR00112590041.

%
\IEEEpeerreviewmaketitle






%

\bibliography{reference}
\bibliographystyle{abbrv}




\section*{Appendix: Detailed Prompt Implementation}

\begin{enumerate}

  \item \textbf{Functionality:} \texttt{Identify critical lines from patches with deleted lines.}

  \textbf{Role:} You are an experienced Linux program analysis expert. 
  I am working on analyzing the Linux kernel patches that fix vulnerabilities. 
  I will give you the content of the patch. You should return the most important 
  and representative lines which are deleted in the patch.

  \textbf{Parameters:} 
  \begin{itemize}
    \item The patch is: \texttt{\$\{patch\_content\}}
    \item The complete functions before the patch are: \texttt{\$\{function\_content\}}
  \end{itemize}

  \textbf{Instruction:} Considering the purpose of the patch, from the lines 
  which are deleted in the patch, pick the \texttt{\$\{num\_lines\}} important 
  and representative lines which are closely related to the logic of the vulnerability.

  \textbf{Output Format:} Output the above most important and representative 
  lines in the format of a Python list \texttt{[]}, each element in the list 
  is a tuple \texttt{(filename, functionname, linenum, line)}. The \texttt{linenum} 
  is the line number inside the corresponding function. When printing the 
  elements, each element is printed in one line instead of multiple lines.

  \item \textbf{Functionality:} \texttt{Identify critical lines from patches with only added lines.}

  \textbf{Role:} You are an experienced Linux program analysis expert. 
  I am working on analyzing the Linux kernel patches that fix vulnerabilities. 
  I will give you the content of the patch and You should return the most important 
  and representative lines.

  \textbf{Parameters:} 
  \begin{itemize}
    \item The patch is: \texttt{\$\{patch\_content\}}
    \item The complete functions before the patch are: \texttt{\$\{function\_content\}}
  \end{itemize}

  \textbf{Instruction:} Considering the purpose of the patch, list important 
  and representative lines with the corresponding functions before the patch 
  is applied. These lines must have data dependency with the added lines 
  in the patch.

  \textbf{Output Format:} Output the above most important and representative 
  lines that exist before the patch in the format of a Python list \texttt{[]}, 
  each element in the list is a tuple \texttt{(filename, functionname, linenum, line)}. 
  The \texttt{linenum} is the line number inside the corresponding function. 
  When printing the elements, each element is printed in one line instead 
  of multiple lines.

  \item \textbf{Functionality:} \texttt{Identify critical lines from patches with only reordered lines.}

  \textbf{Role:} You are an experienced Linux program analysis expert. 
  I am working on analyzing the Linux kernel patches that fix vulnerabilities. 
  I will give you the content of the patch and You should return the most important 
  and representative lines.

  \textbf{Parameters:} 
  \begin{itemize}
    \item The patch is: \texttt{\$\{patch\_content\}}
    \item The complete functions before the patch are: \texttt{\$\{function\_content\}}
    \item The identified reordered lines before the patch are: \texttt{\$\{reorder lines\}}
  \end{itemize}

  \textbf{Instruction:} Considering the purpose of the patch, list important 
  and representative lines with the corresponding functions before the patch 
  is applied. These lines must have data dependency with the reordered lines 
  in the patch.

  \textbf{Output Format:} Output the above most important and representative 
  lines that exist before the patch in the format of a Python list \texttt{[]}, 
  each element in the list is a tuple \texttt{(filename, functionname, linenum, line)}. 
  The \texttt{linenum} is the line number inside the corresponding function. 
  When printing the elements, each element is printed in one line instead 
  of multiple lines.

  \item \textbf{Functionality:} \texttt{Pre-Filtering}

  \textbf{Role:} You are an experienced Linux program analysis expert. 
  I am working on analyzing the Linux kernel patches that fix vulnerabilities. 
  I will give you the content of the patch and the content of a previous commit. 
  You should analyze the patch and understand the logic of the corresponding 
  vulnerability, then determine whether the given commit introduced the vulnerability.

  \textbf{Parameters:}
  \begin{itemize}
    \item The patch is: \texttt{\$\{patch\_content\}}
    \item The content of a previous commit: \texttt{\$\{commit\_content\}}
  \end{itemize}

  \textbf{Instruction:} Analyzing the logic of the patch, determine whether 
  the given commit introduced the vulnerability.

  \textbf{Output Format:} 
  \begin{itemize}
    \item If so, return \texttt{True}, otherwise return \texttt{False}.
    \item If you return \texttt{True}, please also explain the reason why you 
    think the commit introduced the vulnerability.
  \end{itemize}

  \item \textbf{Functionality:} \texttt{Post-Filtering and Result-Finalization}

  \textbf{Role:} You are an experienced Linux program analysis expert. 
  I am working on analyzing the Linux kernel patches that fix vulnerabilities. 
  I will give you the content of the patch (and the corresponding complete 
  function definitions before the patch), also I will provide a list of previous 
  commits (and the corresponding complete function definitions before each commit). 
  You should analyze the patch and understand the logic of the corresponding 
  vulnerability, then determine which commit among the list introduced the vulnerability.

  \textbf{Parameters:}
  \begin{itemize}
    \item The patch is: \texttt{\$\{patch\_content\}}
    \item The below are the lists of previous commits: \texttt{\$\{commit\_content\}}
  \end{itemize}

  \textbf{Instruction:} Analyzing the logic of the patch, determine which 
  commit among the list introduced the vulnerability.

\end{enumerate}




\end{document}